\documentclass[journal]{IEEEtran}
\usepackage{amsmath,amsfonts}
\usepackage{algorithmic}
\usepackage{algorithm}
\usepackage{array}
\usepackage{textcomp}
\usepackage{stfloats}
\usepackage{url}
\usepackage{verbatim}
\usepackage{graphicx}
\usepackage{cite}
\usepackage{subfigure}
\usepackage{wrapfig}
\usepackage{amssymb}
\usepackage{amsthm}
\usepackage{multirow}  
\usepackage{color}
\usepackage{bbding}   
\usepackage{booktabs}       
\usepackage{amsfonts}       
\usepackage{nicefrac}       
\usepackage{microtype}      
\usepackage{xcolor}         
\definecolor{magiccolor}{RGB}{205, 232, 248}
\newtheorem{assumption}{Assumption}[section]
\newtheorem{definition}{Definition}[section]
\newtheorem{theorem}{Theorem}[section]
\newtheorem{proposition}[theorem]{Proposition}

\newcommand{\ours}{{RotoGBML}}

\begin{document}

\title{RotoGBML: Towards Out-of-Distribution Generalization for Gradient-Based Meta-Learning}

\author{Min Zhang$^{1, 2}$, Zifeng Zhuang$^{2}$, Zhitao Wang$^{2}$, Donglin Wang$^{2, \dagger}$, Wenbin Li$^{3}$
\thanks{$^{\dagger}$ Corresponding author, $^{1}$ Zhejiang University, $^{2}$ Westlake University}
\thanks{$^{3}$ Nanjing University, liwenbin@nju.edu.cn}
\thanks{\{zhangmin, zhuangzifeng, wangzhitao, wangdonglin\}@westlake.edu.cn}
}

\markboth{Journal of \LaTeX\ Class Files,~Vol.~14, No.~8, August~2021}%
{Shell \MakeLowercase{\textit{et al.}}: A Sample Article Using IEEEtran.cls for IEEE Journals}


\maketitle

\begin{abstract}
    Gradient-based meta-learning (GBML) algorithms are able to fast adapt to new tasks by transferring the learned meta-knowledge, while assuming that all tasks come from the same distribution (in-distribution, ID). However, in the real world, they often suffer from an out-of-distribution (OOD) generalization problem, where tasks come from different distributions. OOD exacerbates inconsistencies in magnitudes and directions of task gradients, which brings challenges for GBML to optimize the meta-knowledge by minimizing the sum of task gradients in each minibatch. To address this problem, we propose RotoGBML, a novel approach to homogenize OOD task gradients. RotoGBML uses reweighted vectors to dynamically balance diverse magnitudes to a common scale and uses rotation matrixes to rotate conflicting directions close to each other. To reduce overhead, we homogenize gradients with the features rather than the network parameters. On this basis, to avoid the intervention of non-causal features (\textit{e.g.}, backgrounds), we also propose an invariant self-information (ISI) module to extract invariant causal features (\textit{e.g.}, the outlines of objects). Finally, task gradients are homogenized based on these invariant causal features. Experiments show that RotoGBML outperforms other state-of-the-art methods on various few-shot image classification benchmarks.
\end{abstract}

\begin{IEEEkeywords}
    Out-of-distribution, Few-shot learning, Gradient-based meta-learning, Multi-task learning
\end{IEEEkeywords}
\section{Introduction}
\IEEEPARstart{D}{eep} learning has achieved great success in many real-world applications such as visual recognition~\cite{krizhevsky2012imagenet,wang2020deep,srinivas2021} and natural language processing~\cite{vaswani2017attention,devlin2018bert,wu2016google}.
However, deep learning relies heavily on large-scale training data, showing the limitation of not being able to effectively generalize to small data regimes.
To overcome this limitation, researchers have explored and developed a variety of meta-learning algorithms~\cite{vanschoren2019,wang2021meta,RaghuRBV20,thrun2012learning,schmidhuber1987evolutionary}, whose goal is to extract the meta-knowledge over the distribution of tasks rather than instances, and hence compensate for the lack of training data. Among the two dominant strands of meta-learning algorithms, we prefer gradient-based~\cite{FinnAL17,YoonKDKBA18,YinTZLF20,LeeMRS19} over metric-based~\cite{snell2017prototypical,VinyalsBLKW16} for their flexibility and effectiveness. Unfortunately, most of these researches have a restrictive assumption that each task comes from the same distribution (in-distribution, ID). However, distribution shifts among tasks are usually inevitable in real-world scenarios~\cite{shen2021towards,amrith2021two}.
 
In this paper, we consider a realistic scenario, where each minibatch is constructed of tasks from different distributions or datasets (out-of-distribution, OOD). Surprisingly, through repeated experiments, we found that the OOD tasks seriously affect the performance of GBML, \textit{e.g.}, the performance of MAML drops from 75.75\% to 54.29\% with CUB dataset under the 5-way 5-shot setting. 
Intuitively, we explain the possible reason from the optimization objective of the meta-knowledge.
Specifically, GBML algorithms learn the meta-knowledge by minimizing the sum of gradients for each minibatch of tasks (see equation~(\ref{eq:gbml1})).
If task gradients have a significant inconsistency, it may cause the learned meta-knowledge to be dominated by certain tasks with large gradient values and fail to generalize to new tasks, affecting the performance of GBML algorithms.
The OOD generalization problem exacerbates the phenomenon, where task gradients are inconsistent. 
To demonstrate the impact of the OOD problem on task gradients in each minibatch, we next give an illustrative example.

\begin{figure*}[t]
	\centering
	\subfigure[Task space of GBML]{
		\includegraphics[width=3cm,height=3cm]{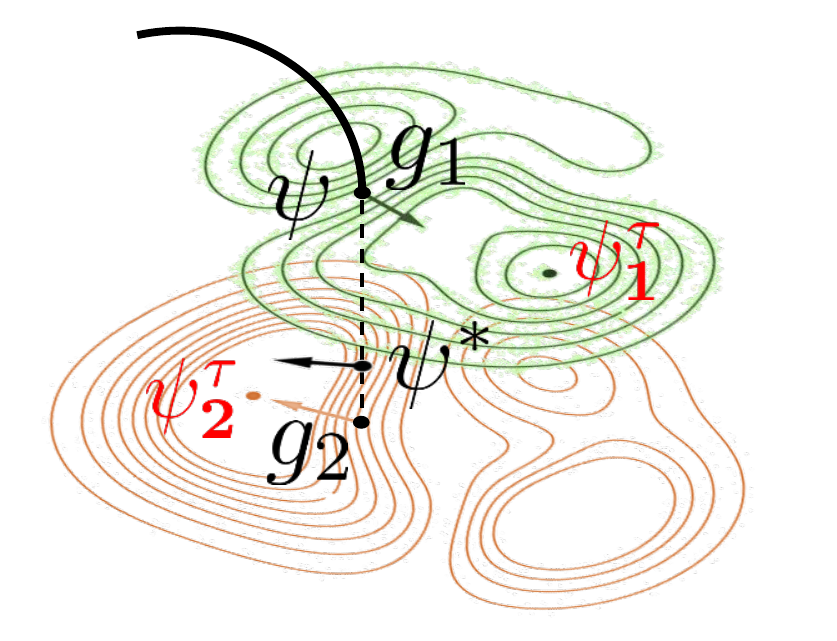}
        \label{fig:frama}}
    \subfigure[Meta space of GBML]{    
		\includegraphics[width=3.3cm,height=3cm]{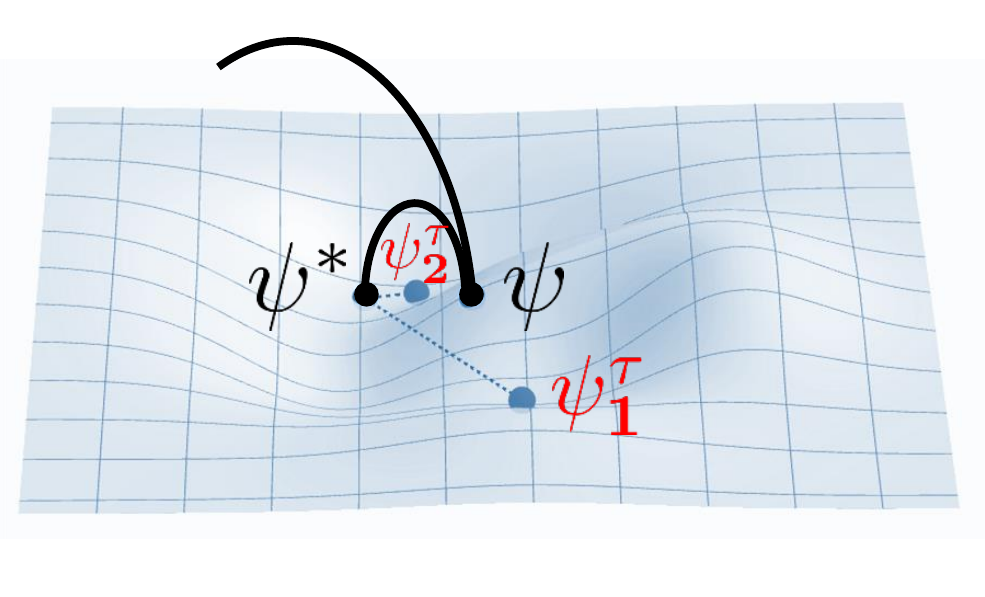}
        \label{fig:framb}}
    \subfigure[Task space of Ours]{      
        \includegraphics[width=3cm,height=3cm]{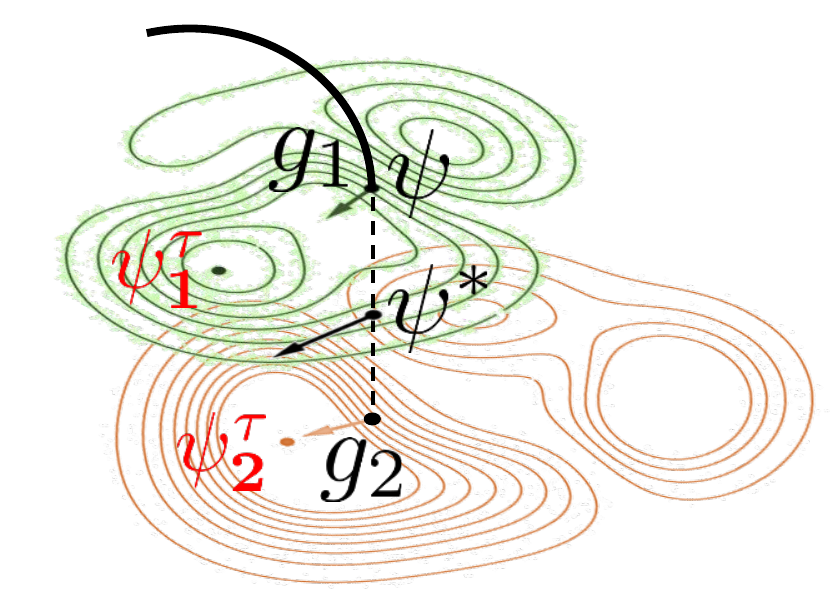}
        \label{fig:framc}}
	\subfigure[Meta space of Ours]{
		\includegraphics[width=3.3cm,height=3cm]{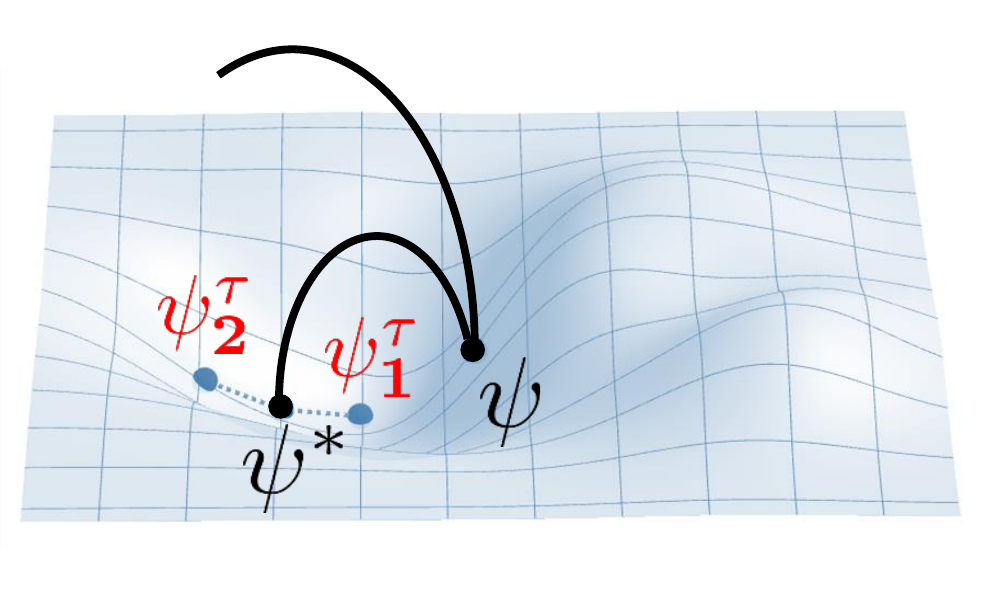}
        \label{fig:framd}}
	\vspace{-2pt}
    \caption{Visualization OOD task-gradient magnitudes and directions for GBML and RotoGBML (Ours) in task and meta spaces. $\psi=(\theta,\phi)$ and $\psi^{*}$ are initial and updated meta parameters, respectively. $g_{1}$ and $g_{2}$ are task gradients with task parameters $\psi_{1}^{\tau}$ and $\psi_{2}^{\tau}$ in outer loop, respectively.}
    \vspace{-4pt}
    \label{fig:fram}
\end{figure*}
 
We take two different distributions to evaluate the impact in Figure~\ref{fig:fram}.
The optimization process of meta-knowledge in GBML has two loops: an inner-loop at task space (Figure~\ref{fig:fram} (a) and (c)) and an outer loop at meta space (Figure~\ref{fig:fram} (b) and (d)). 
We randomly sample two OOD tasks (task1 and task2) from the two different distributions and represent them using green and gray contour lines, respectively.  
First, in the task space, each task learns a task-specific parameter $\psi_{1}^{\tau}$ or $\psi_{2}^{\tau}$ using the same meta-knowledge $\psi$. 
Then, in the meta space, the losses of two OOD tasks are calculated using $\psi_{1}^{\tau}$ and $\psi_{2}^{\tau}$ and are summed in turn to optimize the meta-knowledge from $\psi$ to $\psi^{*}$.
From Figure~\ref{fig:fram} (a), it clearly shows that in the whole optimization process, task2 has a large gradient value (the length of $\|g_{2}\| > \|g_{1}\|$) and the large gradient dominates the learning process, \textit{i.e.}, the updated meta-knowledge $\psi^{*}$ is close to task2 in Figure~\ref{fig:fram} (b).
This causes the learning of meta-knowledge to be dominated by task2 and ignore the existence of task1, and eventually GBML cannot fast adapt to new tasks using the learning meta-knowledge.
When the gradient directions of OOD tasks are conflicting, the meta-knowledge optimized by summing and averaging the two task gradients may counteract each other.    

In this paper, to solve inconsistencies in task-gradient magnitudes and directions, we propose a simple yet effective framework, RotoGBML, to simultaneously homogenize magnitudes and directions and boost the learning of meta-knowledge in GBML.
Specifically, \textcolor{blue}{(1)} RotoGBML solves the gradient magnitudes by dynamically reweighting task gradients at each step of the learning process, while encouraging the learning of ignored tasks. \textcolor{blue}{(2)} Instead of directly modifying gradient directions, RotoGBML smoothly rotates each task space, seamlessly aligning gradient directions in the long run (see Figure~\ref{fig:fram} (c)). 
\textcolor{blue}{(3)} To reduce overhead, we use the features instead of network parameters to homogenize task gradients and more details can be found in Section~\ref{sec:rota}. 
\textcolor{blue}{(4)} We also propose an invariant self-information (ISI) module to extract invariant causal features (\textit{e.g.}, object outline), which are used for homogenization.
The introduction of ISI is mainly because of the fact that the features learned by neural networks are inevitably interfered with by some non-causal features (\textit{e.g.}, image backgrounds), which in turn affects the homogeneity of task gradients.
We theoretically and experimentally demonstrate that the OOD tasks affect meta-knowledge learning process. The main contributions could be briefly summarized as follows:
\begin{itemize}
\item We consider a real-world OOD scenario and propose a general RotoGBML algorithm to solve the OOD generalization problem. RotoGBML helps GBML algorithms learn good meta-knowledge by homogenizing task-gradient magnitudes and directions.
\item To reduce memory, we homogenize task gradients at a feature level. Also, we design an invariant self-information (ISI) module to extract the invariant causal features. Homogenizing gradients using these features provides further guarantees for learning robust meta-knowledge. 
\item We theoretically evaluate our motivation and experimentally demonstrate the effectiveness of RotoGBML algorithm in various few-shot image classification benchmarks.
\end{itemize}
\section{Preliminaries}
\label{sec:pre}

\subsection{Task formulation in GBML.}
In this paper, a supervised learning setting is considered, where each data point is denoted by $(x,y)$ with $x\in X$ being the input and $y\in Y$ being its corresponding label.
GBML assumes $N$ training tasks $\{\mathcal{T}_{i}\}_{i=1}^{N}\sim\mathcal{T}_{tr}$ in each minibatch to be sampled, and an arbitrary testing task $\mathcal{T}_{i}\sim\mathcal{T}_{te}$ is picked.
Formally, each $n$-way $k$-shot task $\mathcal{T}_{i}$ consists of a support set $\mathcal{S}_{i}=(x_{i}^{s}, y_{i}^{s})$ and a query set $\mathcal{Q}_{i}=(x_{i}^{q}, y_{i}^{q})$.
$x_{i}^{s} \in \mathbb{R}^{n_{s} \times d}$, $x_{i}^{q} \in \mathbb{R}^{n_{q} \times d}$, $y_{i}^{s} \in \mathbb{R}^{n_{s} \times n}$ and $y_{i}^{q} \in \mathbb{R}^{n_{q} \times n}$, where $d=c\times h\times w$ indicates the image size, $n_{s}$ and $n_{q}$ denote the number of support and query examples, respectively.
The label spaces of training, validation and testing tasks are different, \textit{i.e.}, $Y_{tr} \cap Y_{val} \cap Y_{te} =\emptyset$.
Most gradient-based meta-learning (GBML) algorithms have a strict assumption for tasks as:
\begin{assumption}
   In-distribution (ID): each task is randomly sampled from the same distribution $\mathbb{P}(\mathcal{T})$, \textit{i.e.}, 
   $\mathcal{T}_{i}\sim\mathbb{P}(\mathcal{T})$ comes from the task set $\{\mathcal{T}_{tr}, \mathcal{T}_{val}, \mathcal{T}_{te}\}$ with $\forall i$.  
 \label{assu:gbml}
\end{assumption}

The assumption indicates a highly restrictive setting due to the following reasons: 
(1) It is contradictory that tasks with disjoint classes come from the same distribution.
(2) The data collection process is susceptible to some unobservable variables, which can cause distribution shifts even from the same dataset.
Improving the generalization of gradient-based meta-learning (GBML) algorithms in the real world is very important.

\subsection{Task optimization in GBML.}
\label{subsec:gbml_ur}
GBML aims to learn a good meta-knowledge and fast adapts to new tasks using two optimization loops (inner loop and outer loop).
Generally, the network architectures of GBML contain a feature encoder $f_{\theta}$ parameterized by $\theta$, which is used to extracted features $f_{\theta}: \mathbb{R}^{d} \rightarrow \mathbb{R}^{m}$, and a classifier $c_{\phi}$ parameterized by $\phi$, which is used to output predict labels $c_{\phi}: \mathbb{R}^{m} \rightarrow \mathbb{R}^{n}$. 
In the inner loop, the model is grounded to an initialization (or meta-knowledge), \textit{i.e.}, $(\phi, \theta)$, which is adapted to the $i$-th task in a few gradient steps $\tau$ (typically 1 $\sim$ 10 steps) by using its support set $\mathcal{S}_{i}$.
In the outer loop, the performance of the adapted model, \textit{i.e.}  $(\phi_{i}^{\tau}, \theta_{i}^{\tau})$, is measured on the query set $\mathcal{Q}_{i}$, and in turn used to optimize the initialization from $(\phi, \theta)$ to $(\phi^{*}, \theta^{*})$.
Let $\mathcal{L}$ denotes the loss function and the above interleaved process is formulated as a bi-level optimization,
\begin{small}
\begin{align}
    \label{eq:gbml1} & (\phi^{*}, \theta^{*}) := \min\limits_{(\phi, \theta)} \frac{1}{N}\sum\nolimits_{i=1}^{N} \mathbb{E}_{(x_{i}^{q}, y_{i}^{q})\sim\mathcal{Q}_{i}} \mathcal{L}_{i}(\phi_{i}^{\tau}(\theta_{i}^{\tau}(x_{i}^{q})), y_{i}^{q}), \\
    \label{eq:gbml2} & \mathrm{s.t.} \ (\phi_{i}^{\tau}, \theta_{i}^{\tau}) \leftarrow \min\limits_{(\phi_{i}^{0}, \theta_{i}^{0})}\mathbb{E}_{(x_{i}^{s}, y_{i}^{s})\sim\mathcal{S}_{i}} \mathcal{L}_{i}(\phi_{i}^{0}(\theta_{i}^{0}(x_{i}^{s})), y_{i}^{s}), 
\end{align}
\end{small}
where equations~(\ref{eq:gbml1}) and~(\ref{eq:gbml2}) are the outer-loop and inner-loop, respectively.
$(\phi_{i}^{0}, \theta_{i}^{0}) = (\phi, \theta)$ is the initial parameters for the $i$-th task.
The learning of meta-knowledge uses the summed and averaged gradients over all tasks in current batch.

\section{Methodology}
\label{sec:method}
To improve the generalization ability of GBML algorithms in real-world scenarios, in this paper, we consider an  out-of-distribution (OOD) setting, \textit{i.e.}, all tasks randomly sampled from a distribution set $\{\mathbb{P}_{i}(\mathcal{T})\}_{i=1}^{N}$. A new assumption on the OOD task distributions is proposed as follows:
\begin{assumption}
   Out-of-distribution (OOD): each task comes from the different distributions $\{\mathbb{P}_{i}(\mathcal{T})\}_{i=1}^{N}$, \textit{i.e.}, $\mathcal{T}_{i} \sim \mathbb{P}_{i}(\mathcal{T})$ and $\mathcal{T}_{j}\sim \mathbb{P}_{j}(\mathcal{T})$ with $ \forall i, j$, $i \neq j$ and $\mathbb{P}_{i}(\mathcal{T}) \neq \mathbb{P}_{j}(\mathcal{T})$.    
   \label{assu:rotogbml}
\end{assumption}

To solve the problem that OOD exacerbates the inconsistencies of task-gradient magnitudes and directions, we introduce RotoGBML, a novel algorithm that consists of two building blocks:
(1) \textit{Reweighting OOD task-gradient magnitudes}. 
A reweighted vector set $\mathbf{w_{\omega}}=\{\omega_{i}\}_{i=1}^{N}$ parameterized by $\omega$ is used to normalize the magnitudes to a common scale (see Section~\ref{sec:reweight});
(2) \textit{Rotating OOD task-gradient directions}. 
A rotation matrix set $R_{\gamma}=\{\gamma_{i}\}_{i=1}^{N}$ parameterized by $\gamma$ is used to rotate the directions close to each other (see Section~\ref{sec:rota}).
The two blocks complement each other and facilitate meta-knowledge to learn common information of all tasks in each minibatch, thereby fast adapting to new tasks with only a few samples. The optimization of RotoGBML is re-defined as:
\begin{small}
\begin{align}
    \label{eq:roto1} & (\phi^{*}, \theta^{*}) := \min\limits_{(\phi,\theta)} \frac{1}{N}\sum\nolimits_{i=1}^{N} \mathbb{E}_{(x_{i}^{q}, y_{i}^{q})\sim\mathcal{Q}_{i}} \textcolor{red}{\omega_{i}}\mathcal{L}_i(\phi_{i}^{\tau}(\textcolor{red}{\gamma_{i}} \theta_{i}^{\tau}(x_{i}^{q})), y_{i}^{q}), \\
    \label{eq:roto2} &\mathrm{s.t.} \ (\phi_{i}^{\tau}, \theta_{i}^{\tau}) \leftarrow \min\limits_{(\phi_{i}^{0}, \theta_{i}^{0})}\mathbb{E}_{(x_{i}^{s}, y_{i}^{s})\sim\mathcal{S}_{i}} \mathcal{L}_{i}(\phi_{i}^{0}(\theta_{i}^{0}(x_{i}^{s})), y_{i}^{s}).
\end{align}
\end{small}
We use \textcolor{red}{red} to annotate the differences between the generic GBML and RotoGBML.
It clearly shows that a set of corresponding $\omega_{i}$ and $\gamma_{i}$ for each task at each batch is used to homogenize gradients in the \textbf{outer loop} optimization process.
And, if task-gradient magnitudes and directions are consistent,  the equation~(\ref{eq:roto1}) degenerates into the equation~(\ref{eq:gbml1}). 
It is worth mentioning that our proposed RotoGBML algorithm is a model-agnostic method that can be equipped with arbitrary GBML algorithms (More details is given in Section~\ref{sec:exper}).

\subsection{Reweighting OOD Task-Gradient Magnitudes}
\label{sec:reweight}
\textbf{How to determine the value of $\mathbf{w_{\omega}}$?} 
This is a key challenge for reweighting OOD task-gradient magnitudes.
Most previous works on multi-task learning use static approaches, such as hyperparameters or prior assignment~\cite{kendall2018multi,crawshaw2020multi}. However, these methods are difficult to adapt to the learning process due to the use of fixed values $\mathbf{w_{\omega}}=\{\omega_{i}\}_{i=1}^{N}$ for each task. 
In Table~\ref{tab:ab}, the experiments of static reweighted vector evaluate the phenomenon. 
Instead, we dynamically adjust $\omega_{i}$ to adapt task distribution shifts by using an optimization strategy over the training iterative process. 
  
Specifically, we initialize $\{\omega_{i}=1|i\in N\}$ for each task in each batch $\mathcal{T}_{b}=\{\mathcal{T}_{i}|i\in N\}$, which aims to treat each task equally at the beginning.
Then, the reweighted vector set is dynamically adjusted by using average gradient norms for the current batch during training.
The optimization objective is as: 
\begin{equation}
    \begin{aligned}
        &\omega^{*} := \min\limits_{\omega} \sum\nolimits_{i=1}^{N} \mathbb{E}_{(x_{i}^{q}, y_{i}^{q})\sim \mathcal{Q}_{i}} \mathcal{L}_{\omega_{i}}(\theta_{i}^{\tau}, x_{i}^{q}, y_{i}^{q}), \\
        &\mathrm{s.t.} \ \mathcal{L}_{\omega_{i}} = \|g_{wi} - \bar{g}_{wb} \times [I_{i}]^{\beta} \|_{1}, 
    \end{aligned}
\label{eq:rewe}
\end{equation}
where $\|\cdot\|_{1}$ is the $\ell_{1}$-norm.
$g_{wi}=\|\nabla_{\theta}\omega_{i}\mathcal{L}_{i}(\theta_{i}^{\tau}(x_{i}^{q}), y_{i}^{q})\|_{2}$ is the $\ell_{2}$-norm of the gradients for each reweighted task loss $\omega_{i}\mathcal{L}_{i}$ in outer loop.
$\bar{g}_{wb}=\frac{1}{N}\sum\nolimits_{i=1}^{N} g_{wi}$ is the average gradient for $\mathcal{T}_{b}$.
In our experimental settings, we only use the gradients of feature encoder $\theta$ to optimize the reweighted vector set $\mathbf{w_{\omega}}$. 
This operation not only saves compute times but also is reasonable because the feature knowledge is more helpfule for the generalization of GBML algorithms~\cite{RaghuRBV20}.
$I_{i}$ aims to balance each task gradient and the formulation is
as follows:
\begin{equation}
    \begin{aligned}
        I_{i} = \hat{\mathcal{L}}_{i}/\sum\nolimits_{i=1}^{N} \hat{\mathcal{L}}_{i}, 
        \ \ \ \mathrm{s.t.} \ \hat{\mathcal{L}}_{i} = \mathcal{L}_{i}/\mathcal{L}_{i}^{0},
    \end{aligned}
\label{eq:ilr}
\end{equation}
where $\mathcal{L}_{i}$ is the cross-entropy loss for query set in outer loop and $\mathcal{L}_{i}^{0}$ is the initial loss to be determineed as $log(n)$ and $n$ is classes.
Specifically, the vaule of $I_{i}$ is used to adjust the tasks of learning too fast or too slow, \textit{i.e.}, the lower value of $I_{i}$ encourages the task to train more slowly, and vice versa. 

In equation~(\ref{eq:rewe}), a novel hyperparameter $\beta$ is introduced to control the learning rate of each task.
When the distributions of tasks very different, a higher $\beta$ should be used to produce the large learning rate.
Conversely, a lower $\beta$ is appropriate for similar tasks.
Note that $\beta=0$ means the learning rate is equal for all tasks.
The dynamically optimized $\omega_{i}$ can be integrated into the learning process to homogenize task-gradient magnitudes. 
In the next section, we describe how to resolve conflicts of task directions in each batch. 

\subsection{Rotating OOD Task-Gradient Directions}
\label{sec:rota}
\textbf{How to define and learn the matrix of $R_{\gamma}$?}
This is a key challege for rotating OOD task-gradient directions.
(1) \textbf{Definition.}
Motivated by previous work~\cite{javaloy2021rotograd}, we initialize $R_{\gamma}\in SO(M)$, where $SO(M)$ is special orthogonal group to denote the set of all rotation matrix $\gamma_{i}$ with size $M$ (the network parameters).
Due to the large size of $M$, to reduce overhead, we focus on feature-level task gradients $\nabla_{\hat{z}_{i}^{q}}\mathcal{L}_{i}$ (rather than $\nabla_{\theta} \mathcal{L}_{i}$).
This is reasonable due to chain rule $\nabla_{\theta}\mathcal{L}_{i} = \nabla_{\theta}\hat{z}_{i}^{q}\cdot \nabla_{\hat{z}_{i}^{q}}\mathcal{L}_{i}$, where $\hat{z}_{i}^{q} = \gamma_{i}z_{i}^{q}$, $z_{i}^{q} = \theta_{i}^{\tau}(x_{i}^{q})$.
Finally, $R_{\gamma}\in SO(m)$ is used, where $m$ is the size of feature dimensions.
(2) \textbf{Learning.}
$R_{\gamma}$ aims to reduce the direction conflict of the task gradients in each batch by rotating the feature space.
We optimize $\gamma_{i}$ to make task spaces closer to each other and the objective is to maximize the cosine similarity or to minimize
\begin{equation}
    \begin{aligned}
        &\gamma^{*} := \min\limits_{\gamma} \sum\nolimits_{i=1}^{N} \mathbb{E}_{(x_{i}^{q}, y_{i}^{q})\sim\mathcal{Q}_{i}}\mathcal{L}_{\gamma_{i}}(\theta_{i}^{\tau}, x_{i}^{q}, y_{i}^{q}),  \\
        &\mathrm{s.t.} \ \mathcal{L}_{\gamma_{i}} = - \left<\gamma_{i}g_{i}, \bar{g}_{rb}\right>, 
    \end{aligned}
    \label{eq:rota}
\end{equation}
where $g_{i}=\nabla_{\theta}\mathcal{L}_i(\theta_{i}^{\tau}(x_{i}^{q}), y_{i}^{q})$, $g_{ri}=\gamma_{i}g_{i}=\nabla_{\hat{z}_{i}^{q}}\mathcal{L}_i(\gamma_{i}\theta_{i}^{\tau}(x_{i}^{q}), y_{i}^{q})$ and $\bar{g}_{rb}=\frac{1}{N}\sum\nolimits_{i\in N}g_{ri}$ is the average gradient for $\mathcal{T}_{b}$.
The optimization of the reweighted vector set in equation~(\ref{eq:rewe}), the rotation matrix set in equation~(\ref{eq:rota}) and the neural network in equation~(\ref{eq:roto1}) can be interpreted as a Stackelberg game: a two player-game in which the leader and follower alternately move to minimize their respective losses, and the leader knows how the followers will react to their moves. 
Such an interpretation allows us to derive simple guidelines to guarantee training convergence, \textit{i.e.}, the network loss does not oscillate as a result of optimizing the two different objectives. We introduce more details in Appendix.

\subsection{Invariant Self-Information}
\label{subsec:isi}
As mentioned above, we homogenize the task gradients from feature level. However, these features learned by neural network are susceptible to some biases, \textit{e.g.}, image backgrounds or textures~\cite{shi2020informative,lecun2015deep,brendel2019approximating}.
For example, in an image of a coat shape with a dog skin texture, the CNN tends to classify it as a dog rather than a coat. This is because the neural network trained with empirical risk minimization (ERM) are more prone to inrobust texture features and ignore robust shape features~\cite{geirhos2018imagenet}. 
In this section, we further introduce the invariant self-information (ISI) module to extract the robust shape features as invariant causal features.
We homogenize OOD task gradients based on these invariant causal features, which helps the reweighted vector set and the rotation matrix set avoid being affected by these biases (\textit{e.g.}, texture features).

\textbf{How to extract the robust shape features?} 
Since there are no existing corresponding shape labels, this is a challenging problem.
Given an image, we observe that the shape regions tend to be more pronounced and carry a high information compared with neighboring regions.
Therefore, at each layer of the neural network, we learn shape features by saving neurons that extract high-informative regions and zeroing out neurons in low-informative regions.
Specifically, we first extract features $z_{i}^{l} \in \mathbb{R}^{c^{l}\times k\times k}$ from input data $x_{i}\in \mathbb{R}^{c\times h\times w}$ in $i$-th task $\mathcal{T}_{i}$, where $l$ is the $l$-th layer of the neural network. Then a drop coefficient $d$ is proposed to drop these less-information regions:
\begin{equation} 
    \begin{aligned}
    &d(z_{i,c}^{l}) \propto e^{-\mathcal{I}(p_{i,k}^{l-1})/T}, \\
    &\mathrm{s.t.} \ \mathcal{I}(p_{i,k}^{l-1}) = -\mathrm{log} \ q_{i,k}^{l-1}(p_{i,k}^{l-1}),
    \end{aligned}
    \label{eq:isi}
\end{equation}
where $p_{i,k}^{l-1}$ is the $k$-th region for $c$-th channel's in $z_{i}^{l-1}$ and $p_{i,k}^{l-1} \in \mathbb{R}^{c_{l-1}\times k\times k}$. $p_{i,k}^{l-1}$ is sampled from the defined distribution $q_{i,k}^{l-1}$. $\mathcal{I}$ denotes self-information and $T$ is temperature. When the amount of information of $p_{i,k}^{l-1}$ is low, that is, the value of this part of the self-information $\mathcal{I}$ is low, the corresponding neuron is not optimized. 

To evaluate $q_{i,k}^{l-1}$, we use Manhattan radius $C$ to obtain $(2C+1)^{2}$ regions as the neighbourhood of $p_{i,k}^{l-1}$. And, we also assume that $p_{i,k}^{l-1}$ and its neighbourhood regions $\mathcal{N}_{i,k}^{l-1}$ come from the same distribution.
We approximate $q_{i,k}^{l-1}(\cdot)$ with its kernel desity estimator and these neighbouring regions, the approximate  representation $\hat{q}_{i,k}^{l-1}$ is shown as follows:
\begin{equation} 
    \hat{q}_{i,k}^{l-1}(p) = \frac{1}{(2C+1)^{2}} \sum\nolimits_{p^{'}\in \mathcal{N}_{i,k}^{l-1}}K(p, p^{'}), 
    \label{eq:qq}
\end{equation}
where $K(\cdot, \cdot)$ is kernel function. 
A Gaussian kernel is used, \textit{i.e.}, $K(p, p^{'})=\frac{1}{\sqrt{2\pi}h}\exp(-\frac{1}{{2h}^{2}}\|p-p^{'}\|^{2})$, where $h$ is the bandwidth.
Then the information of $p_{i,k}^{l-1}$ is estimated by
\begin{equation}
    \hat{\mathcal{I}}(p_{i,k}^{l-1}) = -\mathrm{log} \ \{\sum\nolimits_{p^{'}\in \mathcal{N}_{i,k}^{l-1}}e^{-\frac{1}{{2h}^{2}}\|p_{i,k}^{l-1}-p^{'}\|^{2}}\}. 
\label{eq:final}
\end{equation}

If the more information is included, it can be observed that the larger the difference between $p_{i,k}^{l-1}$ and its neighbourhood regions $\mathcal{N}_{i,k}^{l-1}$. In other words, shapes are more unique in their surroundings and thus more informative.

\subsection{Implementation}
\label{subsec:imp}
We summarize the overall proposed approach in Algorithm~\ref{alg:all}.
In this paper, we consider a more challenging and real-world setting, \textit{i.e.}, tasks are randomly sampled from different distributions during training and testing phases.
\textbf{In the training phase}, we first use the invariant self-information (ISI) module to extract the invariant causal features.
Then, based on these invariant features, we use the reweighted vector set $\mathbf{w_{\omega}}$ to reweight task-gradient magnitudes and the rotating matrix $R_{\gamma}$ to rotate task-gradient directions.
Finally, these homogenized task gradients are used to optimize the meta-knowledge in equation~\ref{eq:roto1}.
\textbf{In the testing phase}, invariant self-information (ISI) module is removed to examine whether our network has truly learned invariant features.
We also remove $\mathbf{w_{\omega}}$ and $R_{\gamma}$ to evaluate the performance of meta-knowledge learned by gradient-based meta-learning (GBML) algorithms.

\begin{algorithm}[t]
	\centering
	\footnotesize
	\caption{GBML Algorithms with RotoGBML}
	\begin{algorithmic}[1]
        \STATE \textit{\, \, \, \, \, \, \, \, \, \, \, \, \, \, --- Meta-Training Phase ---}
        \STATE Randomly initialize all learnable parameters $\{\phi, \theta, \omega, \gamma\}$ 
        \FOR{iteration=1, $\cdots$, MaxIteration}
        \STATE Randomly sample $N$ training tasks of each minibatch from the training distribution set, \textit{i.e.}, $\{\mathcal{T}_{i} = (\mathcal{S}_{i}, \mathcal{Q}_{i}) \}_{i=1}^{N} $ and $\mathcal{T}_{i}\sim\mathcal{T}_{tr}$ with the training distribution set $\{\mathbb{P}_{i}(\mathcal{T})\}_{i=1}^{N}$
        \STATE // \textbf{Inner-loop updated process}
        \FOR{i=1, $\cdots$, MaxBatchTasks=N}
        \STATE Learn invariant causal features using the invariant self-information (ISI) module in Section~\ref{subsec:isi}
        \STATE Compute inner-loop loss $\mathcal{L}_{i}$ and update the parameters from $\{\phi_{i}^{0}, \theta_{i}^{0}\}$ to $\{\phi_{i}^{\tau}, \theta_{i}^{\tau}\}$ using support set
        \STATE Rotate the feature spece of each task in each minibatch from $z_{i}^{q}$ to $\hat{z}_{i}^{q}$ using query set in Section~\ref{sec:rota}
        \STATE Compute outer-loop loss $\mathcal{L}_{i}$ based on the updated model parameters $(\phi_{i}^{\tau}, \theta_{i}^{\tau})$, rotated feature space $\hat{z}_{i}^{q}$ and reweighted loss for each task in each minibatch in Section~\ref{sec:method}
        \ENDFOR
        \STATE // \textbf{Outer-loop updated process}
        \STATE Update main network parameters $(\phi, \theta)$ using equation~(\ref{eq:roto1}) in main paper
        \STATE Update reweighted parameters $\omega$ using equation~(\ref{eq:rewe}) in main paper
        \STATE Update rotation parameters $\gamma$ using equation~(\ref{eq:rota}) in main paper 
        \ENDFOR
        \STATE \textit{\, \, \, \, \, \, \, \, \, \, \, \, \, \, --- Meta-Testing Phase ---} 
        \STATE Randomly sample testing tasks from the testing distribution set, \textit{i.e.}, $\mathcal{T}_{te} \sim \mathbb{P}_{te}(\mathcal{T})$
        \STATE Romove ISI, $\mathbf{w}_{\omega}$ and $R_{\gamma}$ modules and fine-tuning the training model in the inner-loop to adapt new tasks 
        \STATE Predict the performance of fine-tuning model         
    \end{algorithmic}
\label{alg:all}
\end{algorithm}
\section{Theoretical Analysis}
\label{subsec:ga}
In this section, we first theoretically investigate why OOD acerbates inconsistencies in task magnitudes and directions.
We sample two tasks from different distributions, where $\mathcal{T}_{i}=(\mathcal{S}_{i}, \mathcal{Q}_{i})\sim\mathbb{P}_{i}(\mathcal{T}) $ and $\mathcal{T}_{j}=(\mathcal{S}_{j}, \mathcal{Q}_{j})\sim\mathbb{P}_{j}(\mathcal{T})$ with $i \neq j$. In outer loop, the gradient difference is calculated as:
\begin{equation}
    \begin{aligned}
        d_{ij} = &\|\nabla_{(\phi,\theta)} \mathcal{L}_{i}(\phi_{i}^{\tau}(\theta_{i}^{\tau}(x_{i}^{q})), y_{i}^{q}) \\
        &- \nabla_{(\phi,\theta)} \mathcal{L}_{j}(\phi_{j}^{\tau}(\theta_{j}^{\tau}(x_{j}^{q})), y_{j}^{q})\|,
    \end{aligned}
    \label{eq:metric}
\end{equation}
where $g_{i}=\nabla_{(\phi,\theta)} \mathcal{L}_{i}(\phi_{i}^{\tau}(\theta_{i}^{\tau}(x_{i}^{q})), y_{i}^{q})$ and $g_{j}=\nabla_{(\phi,\theta)} \mathcal{L}_{j}(\phi_{j}^{\tau}(\theta_{j}^{\tau}(x_{j}^{q})), y_{j}^{q})$ are gradients of task $\mathcal{T}_{i}$ and $\mathcal{T}_{j}$, respectively.
To bridge the connections of task distributions and task gradients, we introduce \textit{Total Variation Distance (TVD)} to re-estimate $d_{ij}$ on the task distribution.
The definition of Total Variation Distance (TVD) is shown as follows:
\begin{definition}
    (\textit{Total Variation Distance (TVD)}) For two distributions $P$ and $Q$, defined over the sample space $\Omega$ and $\sigma-$field $\mathcal{F}$, the TVD is defined as $\|P-Q\|_{T V}:=\sup _{A \in \mathcal{F}}|P(A)-Q(A)|$.
    \label{eq:tvd}
\end{definition}
It is well-known that the total variation distance (TVD) admits the following characterization
\begin{equation}
    \|P-Q\|_{T V}=\sup _{f: 0 \leq f \leq 1} \mathbb{E}_{x \sim P}[f(x)]-\mathbb{E}_{x \sim Q}[f(x)].
\label{eq:tv}
\end{equation}
\begin{theorem}
    The difference of task gradients $d_{ij}=\|g_{i}-g_{j}\|$ can be bounded by the TVD as: 
    \begin{equation}
        d_{ij} \leq 4 \eta_{base} GL \left\|\mathbb{P}_{i}(\mathcal{T})-\mathbb{P}_{j}(\mathcal{T})\right\|_{TV},
    \end{equation}
\end{theorem}


\begin{figure}[tbp]
    \vspace{-0.45cm}
    \centering
    \includegraphics[width=0.8\linewidth]{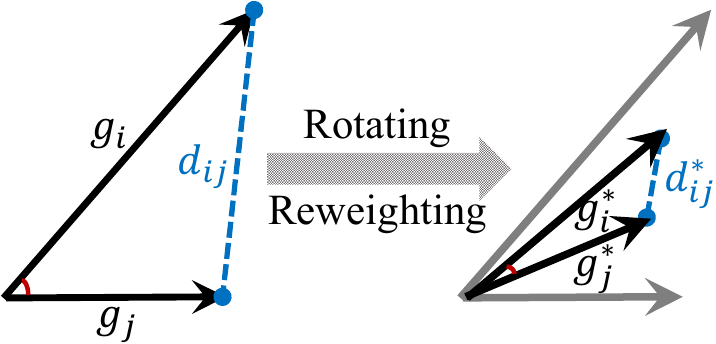}
    \vspace{-0.30cm}
    \caption{The difference of two out-of-distribution (OOD) gradients. (left) The task gradients for GBML algorithms. (right) We short the difference of $d_{ij}^{*}$ by homogenizing the task magnitudes and directions.}
    \vspace{-0.35cm}
    \label{fig:cosine}
\end{figure}

where $\eta_{base}$ is the learning rate of the inner loop. The $G$, $L$ and more proofs are introduced in Appendix. 
When $\mathbb{P}_{i}(\mathcal{T}) \neq \mathbb{P}_{j}(\mathcal{T})$, there is a gradient difference between $\mathcal{T}_{i}$ and $\mathcal{T}_{j}$.
Then, we analyze that RotoGBML can narrow the difference between task gradients from \textit{Cosine Theorem}~\cite{pickover2009math}.
When viewing gradients as a vector, the value of $d_{ij}$ is simplified to calculate the length of the third edge $d_{ij}=\|g_{i}-g_{j}\|$ using \textit{Cosine Theorem} in Figure~\ref{fig:cosine} (left). Based on \textit{Cosine Theorem}, equation~(\ref{eq:metric}) can be re-expressed as $\sqrt{\|g_i\|^2+\|g_j\|^2-2\|g_i\|\|g_j\|\cos \left<g_i, g_j\right>}$, where $\|\cdot\|$ means the length of the vector and $\cos \left<\cdot, \cdot\right>$ is the consine angle between two vectors. 
RotoGBML homogenize task gradients by reweighting task-gradient magnitudes from $\|g_{i}\|$, $\|g_{j}\|$ to $\|g_{i}^{*}\|$, $\|g_{j}^{*}\|$, respectively and rotating task-gradient directions from $\cos \left<g_{i}, g_{j}\right>$ to $\cos \left<g_{i}^{*}, g_{j}^{*}\right>$ to narrow the task gradients difference from $d_{ij}$ to $d_{ij}^{*}$ in Figure~\ref{fig:cosine} (right).
\section{Related Work} 
\label{sec:rela}

\textbf{GBML for few-shot image classification.}
A plethora of GBMLs have been proposed to solve the few-shot image classification problem. 
The goal of these algorithms is to fast adapt to new tasks with only a few samples by transferring the meta-knowledge acquired from training tasks~\cite{FinnAL17, YoonKDKBA18, zintgraf2018caml, flennerhag2021bootstrapped, lee2018gradient}.
However, these works have a strict assumption that training and testing tasks come from the same distribution, which seriously violates the real world.

Recently, some works~\cite{lee2019learning, JeongK20, ChenLKWH19, amrith2021two} have been proposed to study the OOD problem of GBML. However, they mainly focus on the shifts caused by training and testing data (\textit{e.g.}, training from \textit{mini}ImageNet, testing from CUB). 
This setting follows the definition of distribution shifts in traditional machine learning~\cite{shen2021towards}.
They do not consider the problem from the GBML optimization itself and proposed appropriate algorithms for GBML frameworks.
We consider the distribution shifts at the task level due to GBML uses task information of each minibatch to optimize the meta-knowledge.


\textbf{Task gradients homogenization in other fields.}
Homogenization of task-gradient magnitudes and directions is also found in multi-task learning~\cite{javaloy2021rotograd, ChenBLR18, SenerK18, LiuLKXCYLZ21, YuK0LHF20, ChenNHLKCA20}. 
Their goal is to simultaneously learn $K$ different tasks, that is, finding $K$ mappings from a common input dataset to a task-specific set of labels (generally $K$=2).
Motivated by~\cite{javaloy2021rotograd, Wang0L21}, we regard GBML meta-knowledge learning as a multi-task optimization process and homogenize task gradients from a multi-task perspective.
Our work differ from these works in: (1) the input data is different for different tasks in GBML, but the input data of different tasks in multi-task learning is the same. (2) In multi-task learning, existing works on alleviating gradient conflicts usually assume that tasks come from the same distribution. However, we address the problem of conflicting gradients in different distributions. (3) We propose a new invariant self-information module to extract causal features.

\begin{table*}[tbp]
    \vspace{-0.3em}
    \caption{Few-shot image classification average accuracy on the weak OOD generalization problem.}
    \vspace{-4pt}
    \label{tab:weak}
    \begin{center}
    \resizebox{0.7\linewidth}{!}{
    \begin{tabular}{l|cc|cc}
    \toprule
    \multirow{2}{*}{Model}  & \multicolumn{2}{c|}{\textit{mini}ImageNet} & \multicolumn{2}{c}{CUB} \\
    & 5-way 1-shot & 5-way 5-shot & 5-way 1-shot & 5-way 5-shot  \\\midrule
    Linear~\cite{ChenLKWH19} & 42.11\scriptsize$\pm$0.71\% & 62.53\scriptsize$\pm$0.69\% & 47.12\scriptsize$\pm$0.74\% & 64.16\scriptsize$\pm$0.71\%  \\
    ProtoNet~\cite{snell2017prototypical} & 44.42\scriptsize$\pm$0.84\% & 64.24\scriptsize$\pm$0.72\% & 50.46\scriptsize$\pm$0.88\% & 76.39\scriptsize$\pm$0.64\% \\
    MetaOpt~\cite{lee2019meta} & 44.23\scriptsize$\pm$0.59\% & 63.51\scriptsize$\pm$0.48\% & 50.63\scriptsize$\pm$0.67\% & 77.11\scriptsize$\pm$0.59\% \\ 
    MTL~\cite{liu2018meta} & 40.97\scriptsize$\pm$0.54\% & 57.12\scriptsize$\pm$0.68\% & 45.36\scriptsize$\pm$0.75\% & 74.21\scriptsize$\pm$0.79\% \\
    SKD-GEN1~\cite{rajasegaran2020self} & 48.14\scriptsize$\pm$0.45\% & 66.36\scriptsize$\pm$0.36\% & 56.68\scriptsize$\pm$0.67\% & 76.92\scriptsize$\pm$0.91\% \\  \midrule
    ANIL~\cite{RaghuRBV20} & 45.97\scriptsize$\pm$0.32\% & 62.10\scriptsize$\pm$0.45\% & 55.31\scriptsize$\pm$0.63\% & 75.73\scriptsize$\pm$0.64\% \\
    Reptile~\cite{abs-1803-02999} & 45.12\scriptsize$\pm$0.31\% & 58.92\scriptsize$\pm$0.41\% & 56.21\scriptsize$\pm$0.34\% & 72.56\scriptsize$\pm$0.43\% \\ 
    MAML~\cite{FinnAL17} & 46.47\scriptsize$\pm$0.82\% & 62.71\scriptsize$\pm$0.71\% & 54.73\scriptsize$\pm$0.97\% & 75.75\scriptsize$\pm$0.76\%  \\
    iMAML~\cite{RajeswaranFKL19} & 49.30\scriptsize$\pm$1.88\% & 63.24\scriptsize$\pm$0.84\% & 56.14\scriptsize$\pm$0.31\% & 76.05\scriptsize$\pm$0.53\%  \\
    MAML++~\cite{antoniou2018train} & 47.42\scriptsize$\pm$0.32\% & 62.79\scriptsize$\pm$0.52\% & 55.19\scriptsize$\pm$0.34\% & 76.21\scriptsize$\pm$0.54\%  \\ \midrule
    \textbf{ANIL-\ours} & 46.42\scriptsize$\pm$0.33\% & 63.73\scriptsize$\pm$0.42\% & 56.42\scriptsize$\pm$0.32\% & 77.56\scriptsize$\pm$0.45\% \\
    \textbf{Reptile-\ours} & 47.34\scriptsize$\pm$0.31\% & 59.12\scriptsize$\pm$0.51\% & \textcolor{red}{57.87\scriptsize$\pm$0.32\%} & 73.12\scriptsize$\pm$0.53\% \\
    \textbf{MAML-\ours}  & 48.21\scriptsize$\pm$0.31\% & 65.19\scriptsize$\pm$0.42\% & 56.19\scriptsize$\pm$0.32\% & 76.21\scriptsize$\pm$0.45\%   \\
    \textbf{iMAML-\ours} & \textcolor{red}{52.08\scriptsize$\pm$0.31\%} & \textcolor{red}{66.89\scriptsize$\pm$0.47\%} & 57.58\scriptsize$\pm$0.38\% & \textcolor{red}{80.98\scriptsize$\pm$0.44\%}   \\
    \textbf{MAML++-\ours} & 46.40\scriptsize$\pm$0.41\% & 63.08\scriptsize$\pm$0.51\% & 56.80\scriptsize$\pm$0.43\% & 78.72\scriptsize$\pm$0.52\%  \\ \bottomrule
    \end{tabular}}
    \end{center}
    \vspace{-15pt}
  \end{table*} 

\section{Experiments}
\label{sec:exper}

In this section, we conduct comprehensive experiments to evaluate the effectiveness of RotoGBML and compare it with state-of-the-art algorithms.
Specifically, we consider two OOD generalization problems in few-shot image classification: \textit{The weak OOD generalization problem} is training and testing tasks from the same dataset but with disjoint classes~\cite{TriantafillouLZ21}.
\textit{The strong OOD generalization problem} is training and testing tasks from different datasets and has multiple datasets in training data~\cite{amrith2021two, WangD21, TsengLH020}. Unlike these works, we use a more difficult setting, \textbf{\textit{i.e.}, each minibatch of tasks from different datasets}. 

We aim to answer the following questions:
\textbf{Q1:} How does the proposed RotoGBML framework perform for the few-shot image classification task on the weak OOD generalization problem (see Section~\ref{sec:weak})?
\textbf{Q2:} Can RotoGBML fast generalize to new tasks from new distributions when faced with the strong OOD generalization problem (see Section~\ref{sec:strong})? 
\textbf{Q3:} How well each module ($\mathbf{w_{\omega}}$, $R_{\gamma}$ and ISI) in our proposed framework performs in learning the meta-knowledge (see Section~\ref{subsec:ab})?

\noindent \textbf{Datasets.}
For the weak OOD generalization problem, we adopt two popular benchmarks for image classification, \textit{i.e.}, \textit{mini}ImageNet and CUB-200-2011 (\textit{abbr}: CUB). 
For the strong OOD generalization problem, we adopt nine benchmarks, including \textit{mini}ImageNet, CUB, Cars, Places, Plantae, CropDiseases, EuroSAT, ISIC and ChestX following prior works~\cite{TsengLH020, SunLSZCB20, WangD21}.
In the strong OOD generalization, \textit{mini}ImageNet is used only as training data because of its diversity, and evaluate the trained model on the other eight datasets using a leave-out-of method, \textit{i.e.}, randomly sampled one dataset as testing data and other datasets as training data.

\noindent \textbf{Backbone of GBML.}
Since our RotoGBML is model-agnostic and can be equipped with arbitrary GBML algorithms, we use five representative and generic GBML algorithms as our GBML backbone including MAML~\cite{FinnAL17}, MAML++~\cite{antoniou2018train}, ANIL~\cite{RaghuRBV20}, Reptile~\cite{abs-1803-02999}, iMAML~\cite{RajeswaranFKL19}.
In all experimental results, iMAML is used as our GBML backbone due to its good performance, unless otherwise stated. 

\noindent \textbf{Feature encoders.}
We follow previous works~\cite{ChenLKWH19,WangD21} using three general networks as our feature encoder including,
\textit{i.e.} conv4 (filter:64), resnet10 and resnet18.
However, it is well known that GBML algorithms (\textit{e.g}, MAML) under-perform when applied to large networks~\cite{mishra2017simple}, so we use conv4 to show the main experimental results for the weak OOD generalization.
For a fair comparison with some cross-domain methods~\cite{guo2020broader,SunLSZCB20,WangD21,TsengLH020}, we follow their setting using resnet10 for the strong OOD generalization. We also show the performance of the three feature encoders in Figure~\ref{fig:backbone}. 
   
\noindent \textbf{Experimental setting.}
We use the Adam optimizer with the learning rate of inner loop $\eta_{base}=0.01$, outer loop $\eta_{meta}=0.001$ and rotation matrix $\eta_{\gamma}=5e-4$. We set $\beta=0.1$ for weak and $\beta=1.5$ for strong OOD. We optimize all models from scratch and process datasets using data augmentation following previous work~\cite{ChenLKWH19}. We evaluate the performance under two generic settings, \textit{i.e.}, 5-way 1-shot and 5-shot and report the average accuracy as well as 95\% confidence interval.

\subsection{Weak OOD Generalization Problem}
\label{sec:weak}
We conduct experiments on two representative few-shot image classification datasets: \textit{mini}ImageNet and CUB. Table~\ref{tab:weak} reports the experimental results. From Table~\ref{tab:weak}, we have some findings as follows:
\textcolor{blue}{(1)} Compared to Vanilla GBML algorithms (2-nd block), the proposed RotoGBML framework consistently and significantly improves all performances by homogenizing task-gradient magnitudes and directions.
Moreover, compared to the 95\% confidence interval, it is clear that RotoGBML reduces the uncertainty of model predictions and further reduces the high variances of model learning, thereby improving the robustness and generalization.   
\textcolor{blue}{(2)} Compared to some state-of-the-art few-shot image classification methods (1-st block), RotoGBML outperforms all methods, including metric-based model (ProtoNet), fine-tuning model (Linear), pretrain-based model (MTL) and  knowledge-distillation-based model (SKD-GEN1).
This further demonstrates the effectiveness of our proposed RotoGBML algorithm, while also providing a new solution for existing gradient-based meta-learning (GBML) algorithms to outperform metric-based meta-learning methods. 

\begin{figure*}[t]
	\centering
	\includegraphics[width=0.24\textwidth]{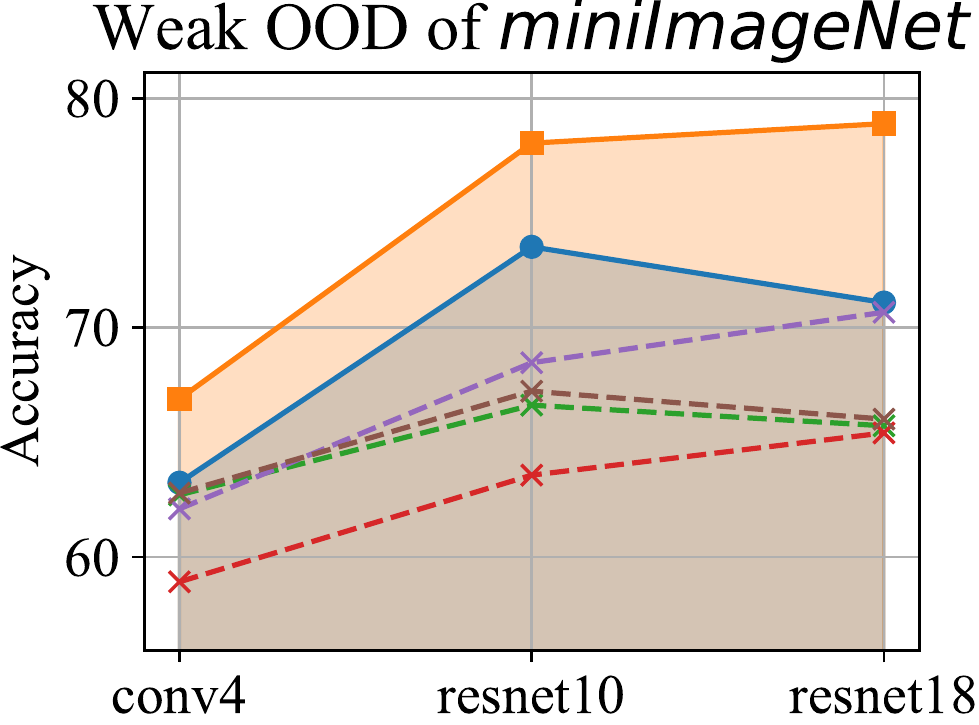} 
  \includegraphics[width=0.24\textwidth]{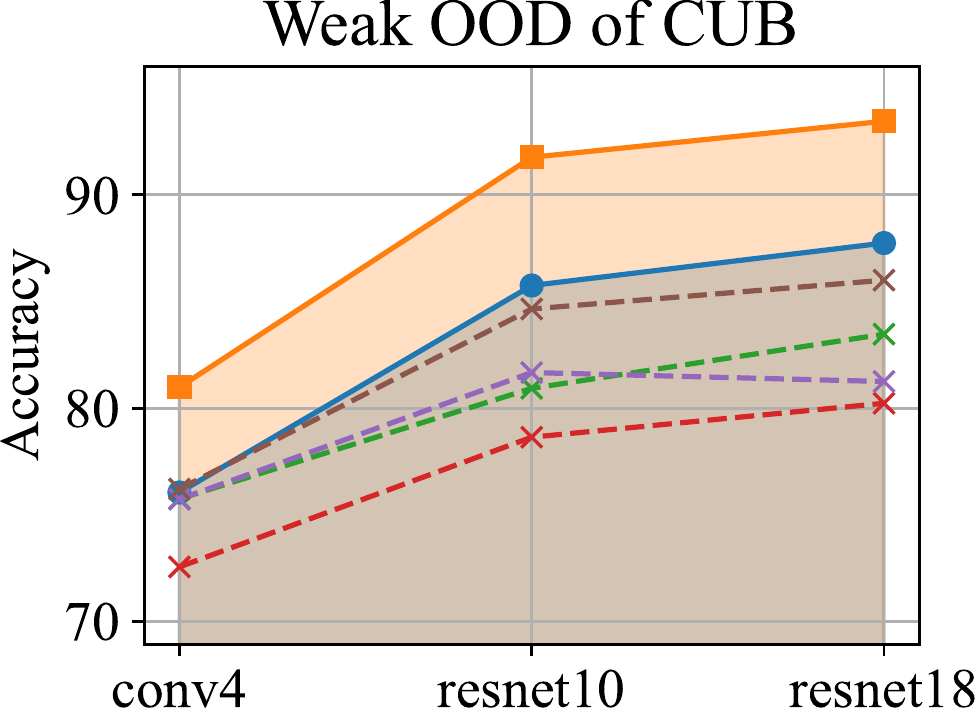}
  \includegraphics[width=0.24\textwidth]{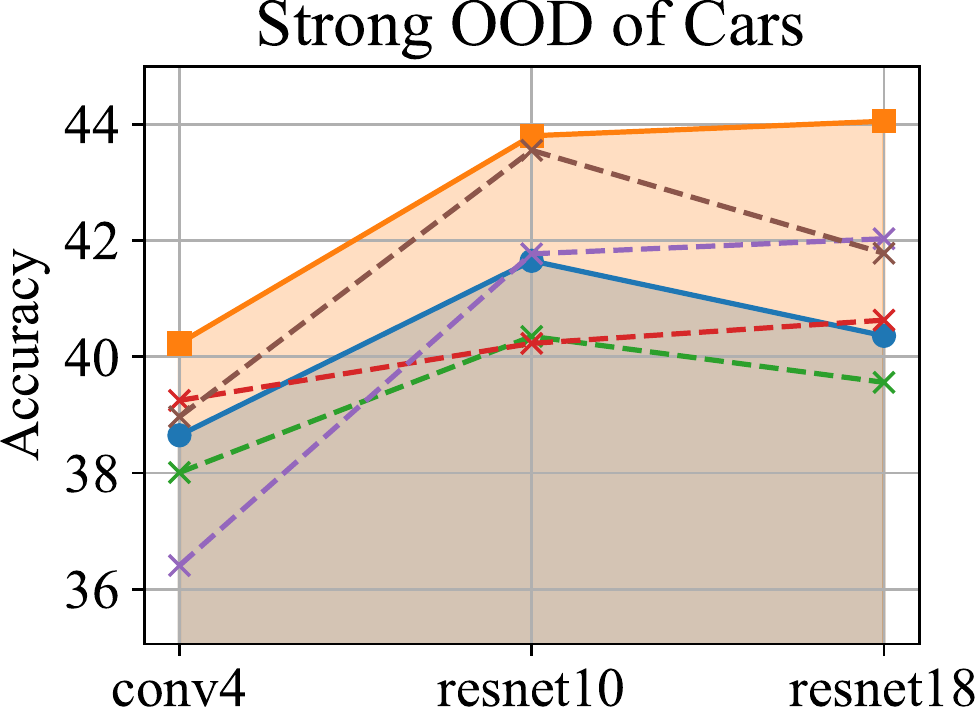}
  \includegraphics[width=0.24\textwidth]{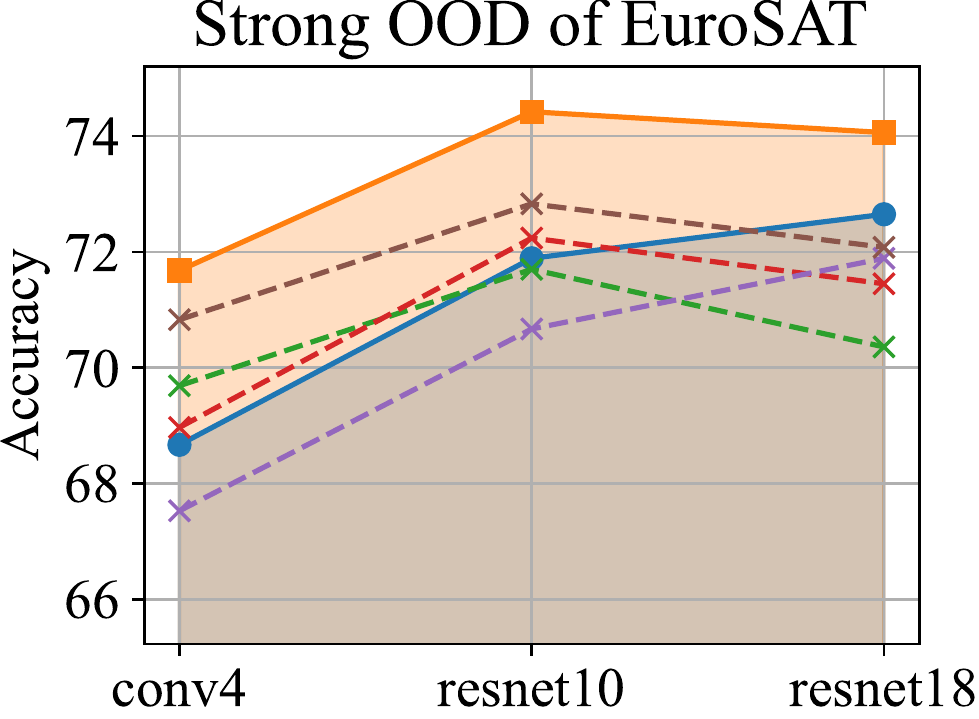} \\
  \includegraphics[width=0.8\textwidth]{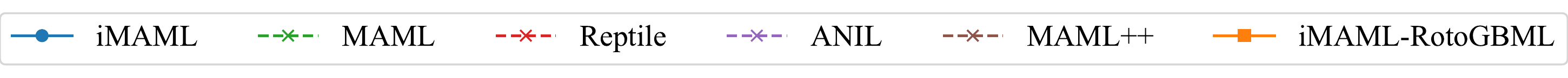}
	\vspace{-1.0em}
  \caption{Few-shot image classification accuracy with different encoders for the weak and strong OOD generalization problems. iMAML-\ours \ has the best performance in all experiments.}
  \vspace{-5mm}
  \label{fig:backbone}
\end{figure*}

We also compare the performance of the model with different feature encoders (conv4, resnet10 and resnet18) in Figure~\ref{fig:backbone}.
We can find that \textcolor{blue}{(1)} our proposed RotoGBML algorithm outperforms all existing gradient-based meta-learning (GBML) algorithms in all experimental settings with different network sizes.
\textcolor{blue}{(2)} It's worth noting that RotoGBML can alleviate the performance degradation problem of large networks.
This is mainly because our proposed ISI module can extract invariant causal features, \textit{i.e.}, the shapes or outlines of objects, which avoid some non-causal features (\textit{e.g.}, the backgrounds, colors or textures of objects) to affect the generalization of large networks on some new tasks from new distributions.

\begin{table*}[htbp]
  \vspace{-0.3em}
  \caption{Few-shot image classification average accuracy on the strong OOD generalization problem.}
  \vspace{-1.0em}
  \begin{center}
    \resizebox{1.0\linewidth}{!}{
      \begin{tabular}{l|cc|cc|cc|cc|cc}
        \toprule
        \multirow{2}{*}{Model} & \multicolumn{2}{c|}{CUB} & \multicolumn{2}{c|}{Cars} & \multicolumn{2}{c|}{Places} & \multicolumn{2}{c|}{Plantae} & \multicolumn{2}{c}{Average} \\
        & 1-shot & 5-shot & 1-shot & 5-shot & 1-shot & 5-shot & 1-shot & 5-shot & 1-shot & 5-shot \\\midrule
        Random~\cite{guo2020broader}        & 40.53\% & 53.76\% & 28.12\% & 39.21\% & 47.57\% & 61.68\% & 30.77\% & 40.45\% & 36.75\% & 48.78\% \\
        RelationNet~\cite{sung2018learning} & 41.27\% & 56.77\% & 30.09\% & 40.46\% & 48.16\% & 64.25\% & 31.23\% & 42.71\% & 37.69\% & 51.05\% \\
        RelationNet-FT~\cite{TsengLH020}    & 43.33\% & 59.77\% & 30.45\% & 40.18\% & 49.92\% & 65.55\% & 32.57\% & 44.29\% & 39.07\% & 52.45\% \\
        RelationNet-LRP~\cite{SunLSZCB20}   & 41.57\% & 57.70\% & 30.48\% & 41.21\% & 48.47\% & 65.35\% & 32.11\% & 43.70\% & 38.16\% & 51.99\% \\
        RelationNet-ATA~\cite{WangD21}      & 43.02\% & 59.36\% & 31.79\% & 42.95\% & 51.16\% & 66.90\% & 33.72\% & 45.32\% & 39.92\% & 53.63\% \\ \midrule
        MAML~\cite{FinnAL17}                & 40.66\% & 54.29\% & 30.02\% & 40.35\% & 45.93\% & 60.00\% & 31.35\% & 44.65\% & 36.99\% & 49.82\% \\
        iMAML~\cite{RajeswaranFKL19}        & 43.56\% & 57.98\% & 31.25\% & 41.65\% & 46.14\% & 61.52\% & 32.01\% & 43.65\% & 38.24\% & 51.20\% \\
        MAML++~\cite{antoniou2018train}     & 42.11\% & 58.72\% & 32.93\% & 43.55\% & 46.52\% & 62.61\% & 33.91\% & 42.77\% & 38.87\% & 51.91\% \\ 
        ANIL~\cite{RaghuRBV20}              & 42.67\% & 55.58\% & 30.63\% & 41.77\% & 45.55\% & 61.72\% & 31.90\% & 45.95\% & 37.69\% & 51.26\% \\   \midrule 
        \textbf{MAML-\ours}                 & 41.38\% & 56.82\% & \textcolor{red}{39.19\%} & 42.35\% & 46.12\% & 62.94\% & 34.21\% & 46.21\% & 40.23\% & 52.08\% \\
        \textbf{iMAML-\ours} & \textcolor{red}{46.12\%} & \textcolor{red}{60.23\%} & 35.81\% & \textcolor{red}{43.80\%} & \textcolor{red}{52.65\%} & 67.47\% & \textcolor{red}{35.24\%} & 46.51\% & \textcolor{red}{42.46\%} & \textcolor{red}{54.50\%}   \\
        \textbf{MAML++-\ours} & 44.38\% & 59.66\% & 33.66\% & 42.23\% & 50.14\% & \textcolor{red}{68.77\%} & 34.14\% & 45.23\% & 40.58\% & 53.97\%  \\
        \textbf{ANIL-\ours} & 45.86\% & 58.81\% & 31.68\% & 42.01\% & 47.20\% & 62.44\% & 32.38\% & \textcolor{red}{46.91\%} & 39.28\% & 52.54\%    \\
        \midrule
        \multirow{2}{*}{Model} & \multicolumn{2}{c|}{ChestX} & \multicolumn{2}{c|}{CropDiseases} & \multicolumn{2}{c|}{EuroSAT} & \multicolumn{2}{c|}{ISIC} & \multicolumn{2}{c}{Average} \\
        & 1-shot & 5-shot & 1-shot & 5-shot & 1-shot & 5-shot & 1-shot & 5-shot & 1-shot & 5-shot \\\midrule
        Random~\cite{guo2020broader}        & 19.81\% & 21.80\% & 50.43\% & 69.68\% & 40.97\% & 58.00\% & 28.56\% & 37.91\% & 34.94\% & 46.85\% \\
        RelationNet~\cite{sung2018learning} & 21.95\% & 24.07\% & 53.58\% & 72.86\% & 49.08\% & 65.56\% & 30.53\% & 38.60\% & 38.79\% & 50.27\% \\
        RelationNet-FT~\cite{TsengLH020}    & 21.79\% & 23.95\% & 57.57\% & 75.78\% & 53.53\% & 69.13\% & 30.38\% & 38.68\% & 40.82\% & 51.89\% \\
        RelationNet-LRP~\cite{SunLSZCB20}   & 22.11\% & 24.28\% & 55.01\% & 74.21\% & 50.99\% & 67.54\% & 31.16\% & 39.97\% & 39.82\% & 51.50\% \\
        RelationNet-ATA~\cite{WangD21} & 22.14\% & 24.43\% & \textcolor{red}{61.17\%} & 78.20\% & 55.69\% & 71.02\% & 31.13\% & 40.38\% & \textcolor{red}{42.53\%} & 53.50\% \\ \midrule
        MAML~\cite{FinnAL17}                & 21.72\% & 23.48\% & 52.54\% & 78.05\% & 48.10\% & 71.70\% & 28.58\% & 40.13\% & 37.74\% & 53.34\% \\
        iMAML~\cite{RajeswaranFKL19}        & 22.45\% & 24.14\% & 53.23\% & 79.23\% & 50.34\% & 71.89\% & 29.58\% & 45.65\% & 38.90\% & 55.23\% \\
        MAML++~\cite{antoniou2018train}     & 21.43\% & 22.67\% & 52.70\% & 76.45\% & 46.25\% & 72.83\% & 30.84\% & 44.51\% & 37.81\% & 54.12\% \\  
        ANIL~\cite{RaghuRBV20}              & 21.36\% & 21.83\% & 51.62\% & 77.30\% & 49.81\% & 70.67\% & 30.04\% & 45.30\% & 38.21\% & 53.78\% \\
        \midrule
        \textbf{MAML-\ours}                 & 22.92\% & 24.51\% & 57.72\% & 80.17\% & 51.93\% & 72.48\% & 30.94\% & 43.89\% & 40.88\% & 55.26\% \\
        \textbf{iMAML-\ours} & \textcolor{red}{24.12\%} & 25.98\% & 58.34\% & \textcolor{red}{80.23\%} & \textcolor{red}{55.78\%} & \textcolor{red}{74.42\%} & 31.25\% & \textcolor{red}{49.68\%} & 42.37\% & \textcolor{red}{57.58\%}  \\
        \textbf{MAML++-\ours}               & 22.97\% & 23.26\% & 55.25\% & 78.51\% & 48.44\% & 73.57\% & \textcolor{red}{32.48\%} & 47.53\% & 39.79\% & 55.72\% \\
        \textbf{ANIL-\ours}                 & 23.55\% & \textcolor{red}{26.47\%} & 54.88\% & 79.50\% & 50.21\% & 71.22\% & 31.21\% & 46.90\% & 39.96\% & 56.02\%  \\
        \bottomrule
      \end{tabular}}
  \end{center}
  \vspace{-15pt}
\label{tab:strong}
\end{table*}

\begin{table*}[t]
  \caption{Ablation study on all modules of RotoGBML framework. The first row represents the performance of iMAML. \textcolor{violet}{Violet} is the weak OOD and \textcolor{blue}{blue} is the strong OOD experiments. \textit{hyper} means that a fixed reweighted vector set is used to reweight each task gradients in each minibatch.}
  \vspace{-10pt}
  \label{tab:ab}
  \begin{center}
  \resizebox{1.0\linewidth}{!}{
  \begin{tabular}{ccc|cc|cc|cc|cc}
  \toprule
  \multirow{2}{*}{$R_{\gamma}$} & \multirow{2}{*}{$\mathbf{w_{\omega}}$} & \multirow{2}{*}{ISI} & \multicolumn{2}{c|}{\textit{mini}ImageNet} & \multicolumn{2}{c|}{CUB} & \multicolumn{2}{c|}{Cars} & \multicolumn{2}{c}{EuroSAT} \\
  & & & 1-shot & 5-shot & 1-shot & 5-shot & 1-shot & 5-shot & 1-shot & 5-shot   \\ \midrule
  {\footnotesize\XSolidBrush} & {\footnotesize\XSolidBrush}  & {\footnotesize\XSolidBrush} & 49.30\% & 63.24\% & 56.14\% & 76.05\% & 31.25\% & 41.65\% & 50.34\% & 71.89\% \\ \midrule
  {\footnotesize\XSolidBrush} & {\footnotesize\textit{hyper}}  & {\footnotesize\XSolidBrush} & 48.25\%(\textcolor{violet}{-1.1}) & 59.45\%(\textcolor{violet}{-3.8}) & 56.78\%(\textcolor{violet}{+0.6}) & 77.02\%(\textcolor{violet}{+1.0}) & 30.14\%(\textcolor{blue}{-1.1}) & 38.45\%(\textcolor{blue}{-3.2}) & 48.46\%(\textcolor{blue}{-1.9}) & 70.58\%(\textcolor{blue}{-1.3}) \\
  {\footnotesize\XSolidBrush} & {\footnotesize\CheckmarkBold} & {\footnotesize\XSolidBrush} & 50.45\%(\textcolor{violet}{+1.2}) & 64.77\%(\textcolor{violet}{+1.5}) & 56.98\%(\textcolor{violet}{+0.8}) & 78.39\%(\textcolor{violet}{+2.3}) & 33.36\%(\textcolor{blue}{+2.1}) & 43.08\%(\textcolor{blue}{+1.4}) & 53.81\%(\textcolor{blue}{+3.5}) & 72.65\%(\textcolor{blue}{+0.8})    \\
  {\footnotesize\CheckmarkBold} & {\footnotesize\XSolidBrush} & {\footnotesize\XSolidBrush} & 50.42\%(\textcolor{violet}{+1.1}) & 64.53\%(\textcolor{violet}{+1.3}) & 57.16\%(\textcolor{violet}{+1.0}) & 78.61\%(\textcolor{violet}{+2.6}) & 33.59\%(\textcolor{blue}{+2.3}) & 43.36\%(\textcolor{blue}{+1.7}) & 53.62\%(\textcolor{blue}{+3.3}) & 73.12\%(\textcolor{blue}{+1.2}) \\
  {\footnotesize\XSolidBrush} & {\footnotesize\XSolidBrush} & {\footnotesize\CheckmarkBold} & 51.56\%(\textcolor{violet}{+2.3}) & 65.17\%(\textcolor{violet}{+2.3}) & 57.05\%(\textcolor{violet}{+0.9}) & 77.69\%(\textcolor{violet}{+1.6}) & 31.69\%(\textcolor{blue}{+0.9}) & 40.04\%(\textcolor{blue}{+0.6}) & 51.18\%(\textcolor{blue}{+0.8}) & 72.43\%(\textcolor{blue}{+0.5}) \\ \midrule
  {\footnotesize\CheckmarkBold} & {\footnotesize\CheckmarkBold} & {\footnotesize\XSolidBrush} & 50.84\%(\textcolor{violet}{+1.5}) & 65.37\%(\textcolor{violet}{+2.1}) & 57.47\%(\textcolor{violet}{+1.3}) & 79.11\%(\textcolor{violet}{+3.1}) & 34.41\%(\textcolor{blue}{+3.2}) & 43.59\%(\textcolor{blue}{+1.9}) & 54.21\%(\textcolor{blue}{+3.9}) & 73.80\%(\textcolor{blue}{+1.9})  \\
  {\footnotesize\XSolidBrush} & {\footnotesize\CheckmarkBold} & {\footnotesize\CheckmarkBold} & 51.81\%(\textcolor{violet}{+2.5}) & 66.61\%(\textcolor{violet}{+3.4}) & 57.38\%(\textcolor{violet}{+1.2}) & 80.56\%(\textcolor{violet}{+4.5}) & 33.88\%(\textcolor{blue}{+2.6}) & 43.18\%(\textcolor{blue}{+1.5}) & 54.49\%(\textcolor{blue}{+4.2}) & 73.43\%(\textcolor{blue}{+1.5})   \\
  {\footnotesize\CheckmarkBold} & {\footnotesize\XSolidBrush} & {\footnotesize\CheckmarkBold} & 51.95\%(\textcolor{violet}{+2.7}) & 66.74\%(\textcolor{violet}{+3.5}) & 57.23\%(\textcolor{violet}{+1.1}) & 80.60\%(\textcolor{violet}{+4.6}) & 35.01\%(\textcolor{blue}{+3.8}) & 43.47\%(\textcolor{blue}{+1.8}) & 54.85\%(\textcolor{blue}{+4.5}) & 73.95\%(\textcolor{blue}{+2.1})  \\ \midrule
  {\footnotesize\CheckmarkBold} & {\footnotesize\CheckmarkBold} & {\footnotesize\CheckmarkBold} & 52.08\%(\textcolor{violet}{+2.8}) & 66.89\%(\textcolor{violet}{+3.7}) & 57.58\%(\textcolor{violet}{+1.4}) & 80.98\%(\textcolor{violet}{+4.9}) & 35.81\%(\textcolor{blue}{+4.6}) & 43.80\%(\textcolor{blue}{+2.1}) & 55.78\%(\textcolor{blue}{+5.4}) & 74.42\%(\textcolor{blue}{+2.5})  \\
  \bottomrule
  \end{tabular}}
  \end{center}
  \vspace{-8pt}
\end{table*} 

\subsection{Strong OOD Generalization Problem}
\label{sec:strong}
In this section, we consider a more challenging OOD setting, \textit{i.e.}, each task of each minibatch (typically the number of tasks $N=4$) comes from different datasets.
We compare our method and some state-of-the-art methods with the strong OOD generalization problem in Table~\ref{tab:strong}.
From Table~\ref{tab:strong}, we have some significant findings as follows:
\textcolor{blue}{(1)} Compared to Vanilla GBML algorithms (2-nd block), RotoGBML outperforms all algorithms on eight datasets with different performance gains.
These results again prove that our RotoGBML can learn robust meta-knowledge, which in turn generalizes quickly to new tasks from new distributions.
\textcolor{blue}{(2)} Compared to some cross-domain methods (1-st block), RotoGBML has a good performance in most settings, which demonstrates that we can help GBML algorithms to achieve state of the art by homogenizing task-gradient magnitudes and directions.
\textcolor{blue}{(3)} The improvement of the strong OOD setting is usually larger than that of the weak OOD setting (Table~\ref{tab:weak}).
The phenomenon justifies our conclusion that the larger the distribution difference, the more obvious inconsistencies in magnitudes and directions of task gradients.

In Figure~\ref{fig:backbone}, the performances of the strong OOD problem are shown with different network sizes.
We show the results on Cars and EuroSAT as we
have similar observations on other datasets.
Compared to Vanilla GBML algorithms, iMAML has the best performance on the weak OOD problem, but significantly degrades on the strong OOD problem.
Because it uses a regularization to encourage task parameters to close to meta parameters, large distribution shifts affect the learning of meta parameters.
Our RotoGBML can learn good meta parameters by homogenizing task gradients.

\subsection{Ablation Study}
\label{subsec:ab}
\begin{figure*}[t]
	\centering
    \includegraphics[width=1.0\textwidth]{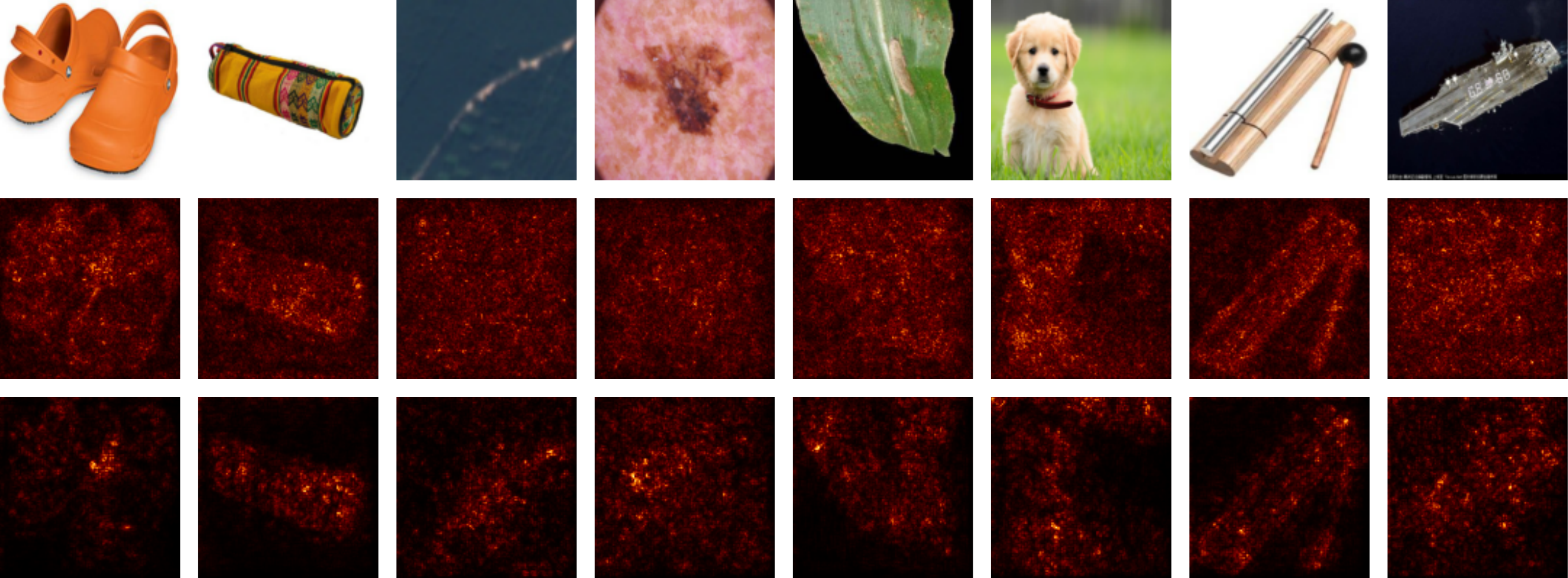}
	\vspace{-15pt}
  \caption{Visualization of resnet10's saliency (gradient) map. For each subfigure, from top to bottom: original image, saliency map of vanilla iMAML and saliency map of iMAML with our ISI module.}
  \label{fig:sm}
\end{figure*}

\paragraph{The performance of our proposed three modules}
We use a hierarchical ablation study to evaluate the performance of each module for RotoGBML under the weak and strong OOD problems in Table~\ref{tab:ab}.
The 1-st row represents the results of Vanilla iMAML.
The plus sign means performance increase and the minus sign means performance decrease, where \textcolor{violet}{violet} is the weak OOD and \textcolor{blue}{blue} is the strong OOD experiments. 
To evaluate the performances of a fixed reweighted vector set, \textit{e.g.}, [0.1,0.5,0.3,0.4] is used for the four tasks in each minibatch in our experiments, the results are shown in the 2-nd row.
From Table~\ref{tab:ab}, the following findings:
\textcolor{blue}{(1)} compared to Vanilla iMAML, the performances of hyperparameters have a significant drop in most settings, but our dynamic adjustment strategy (3-rd row) has a clear improvement.
This further evaluates the effectiveness of our learning method to dynamically optimize the reweighted vectors.
\textcolor{blue}{(2)} Each module has different performance gains (3$\sim$5-th rows). Compared to the strong OOD problem, the performances of ISI on the weak OOD problem have a large improvement, because even if invariant features are learned, the strong inconsistencies of task gradients affect the learning of meta-knowledge.
\textcolor{blue}{(3)} Arbitrary combination of two modules has good performance in 6$\sim$8-th rows, especially with ISI. This is because adjusting the gradients based on invariant robust information from different distributions can learn better. We also give some visualization experiments of ISI learning invariant features in Figure~\ref{fig:sm}.
And all modules are used to have the best performance.

\begin{figure*}[t]
	\centering
	\includegraphics[width=0.29\textwidth]{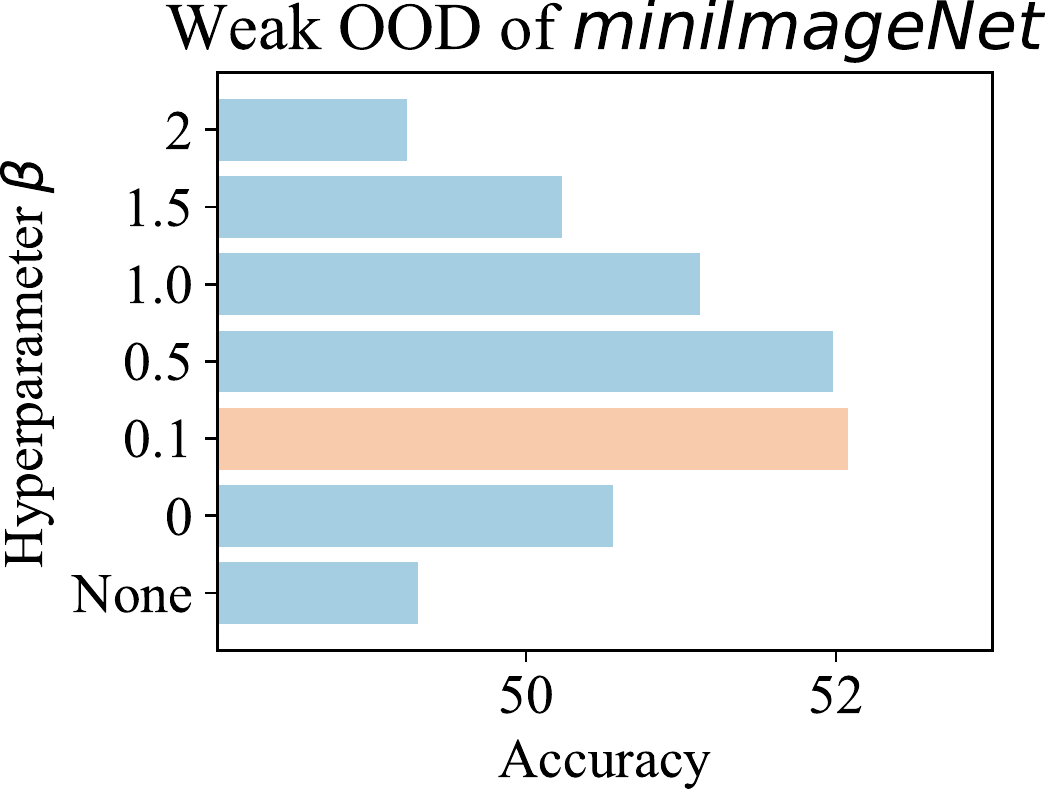} 
    \includegraphics[width=0.22\textwidth]{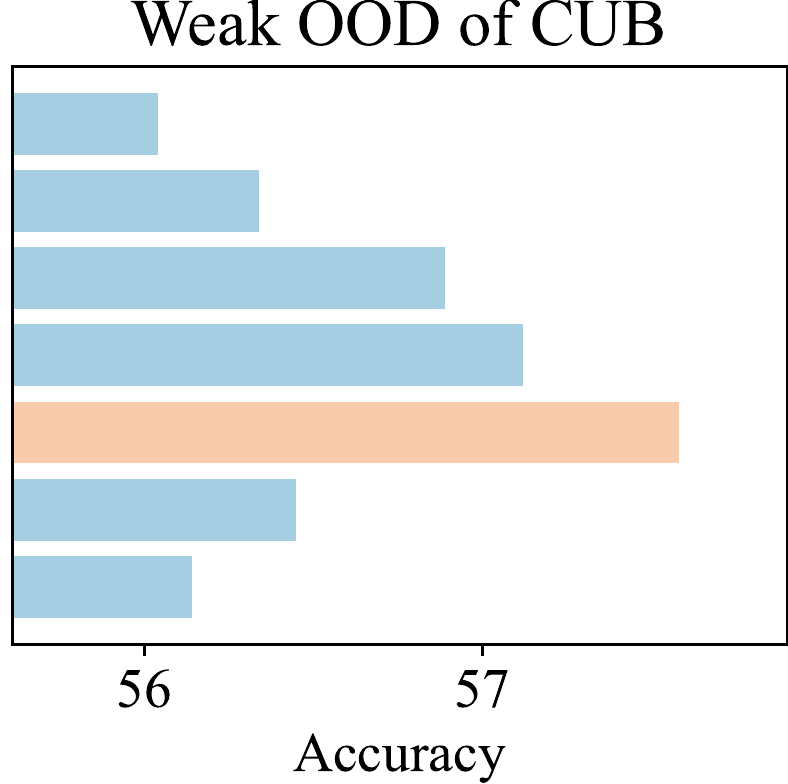}
    \includegraphics[width=0.22\textwidth]{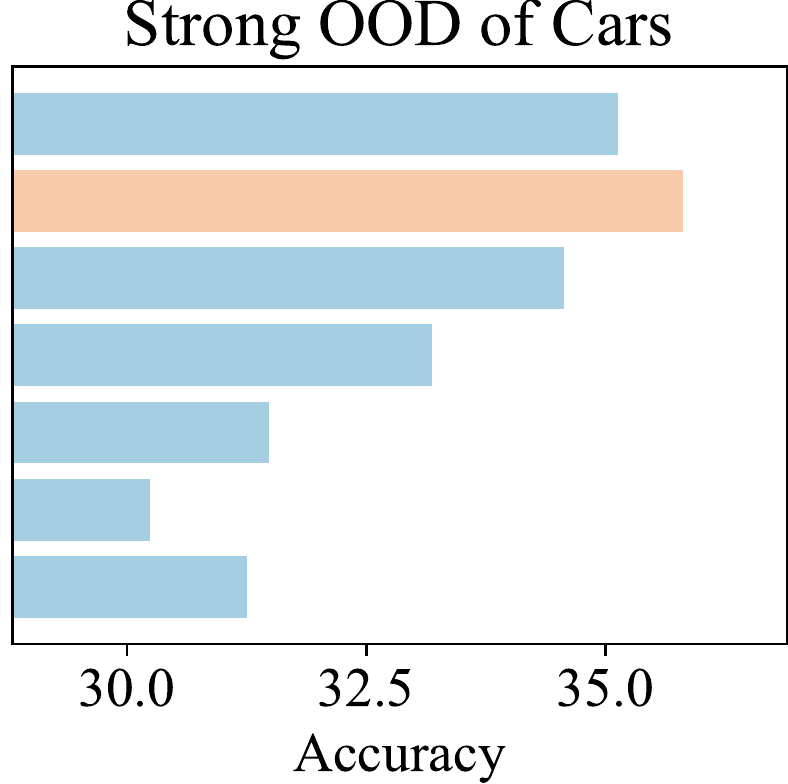}
    \includegraphics[width=0.22\textwidth]{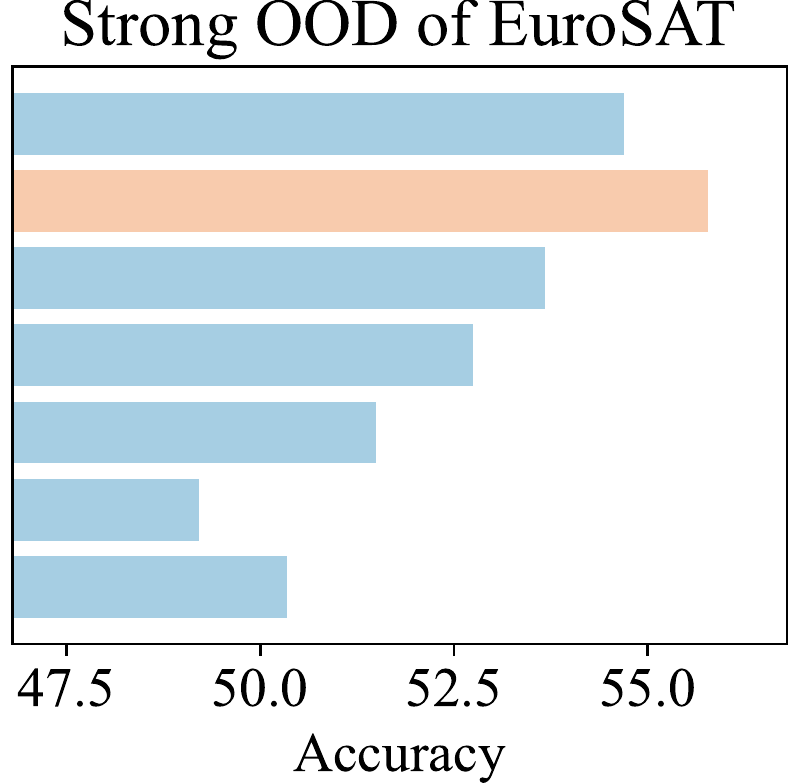} 
	\vspace{-4pt}
  \caption{Few-shot classification accuracy with different hyperparameter $\beta$ for the weak and strong OOD generalization problems under the 5-way 1-shot setting.}
  \label{fig:abeta1}
  \vspace{-5pt}
\end{figure*}

\begin{figure*}[t]
	\centering
	\includegraphics[width=0.29\textwidth]{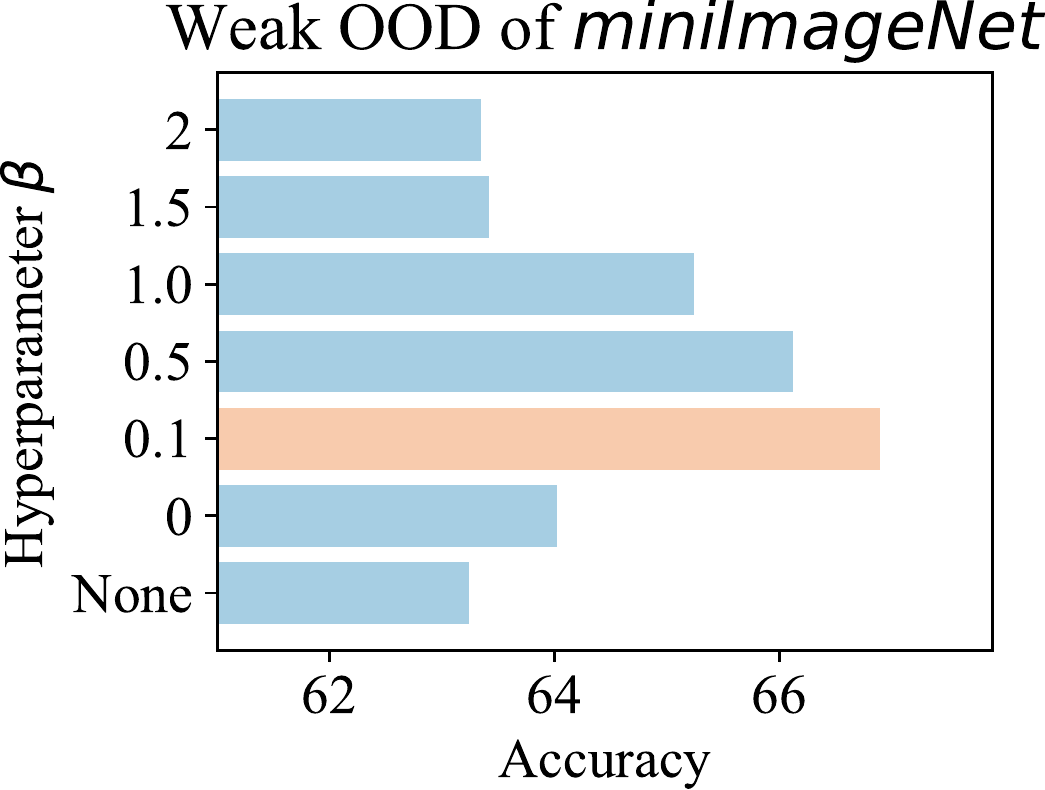} 
    \includegraphics[width=0.22\textwidth]{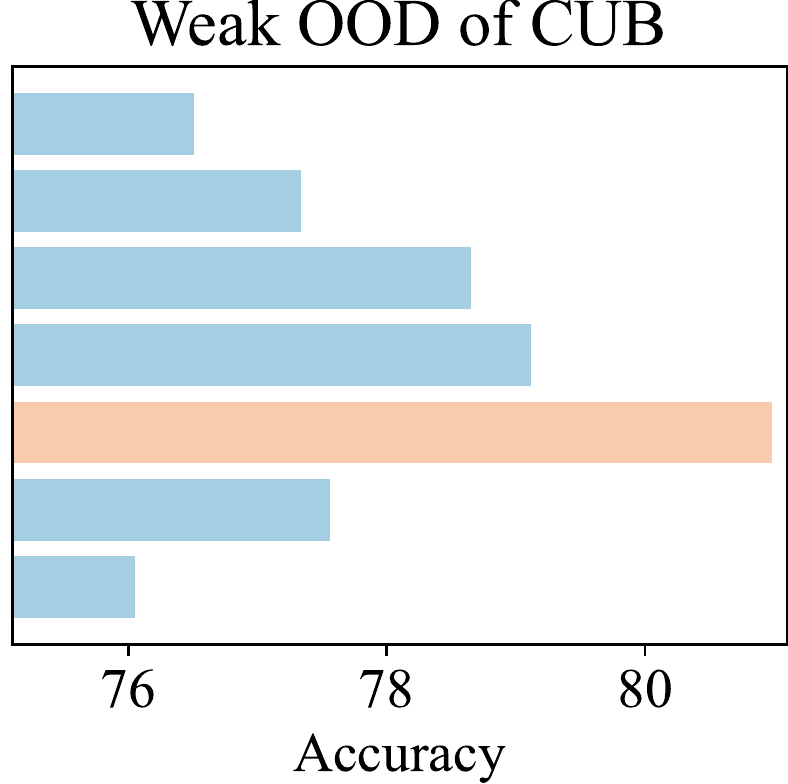}
    \includegraphics[width=0.22\textwidth]{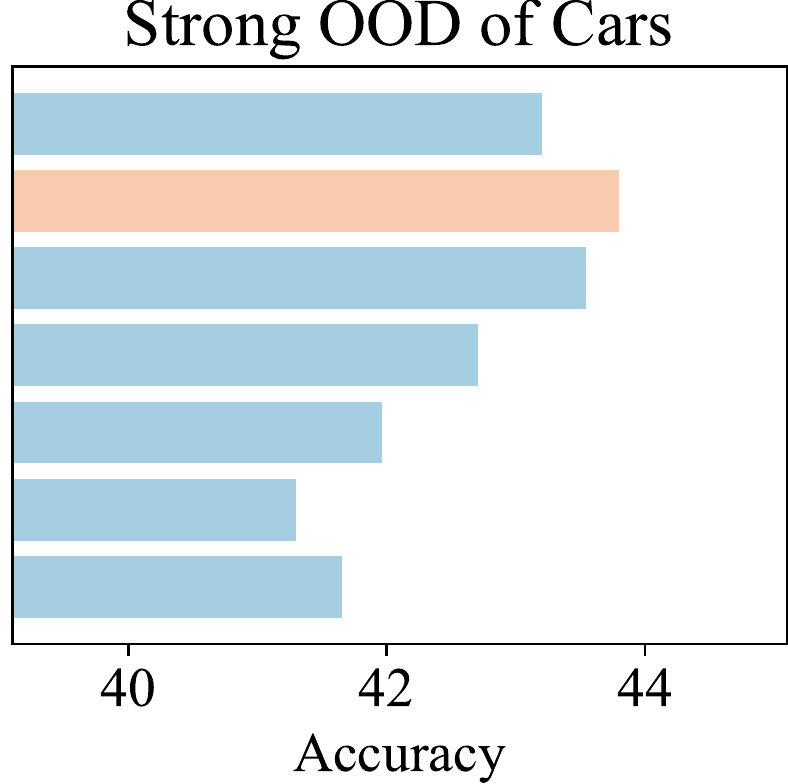}
    \includegraphics[width=0.22\textwidth]{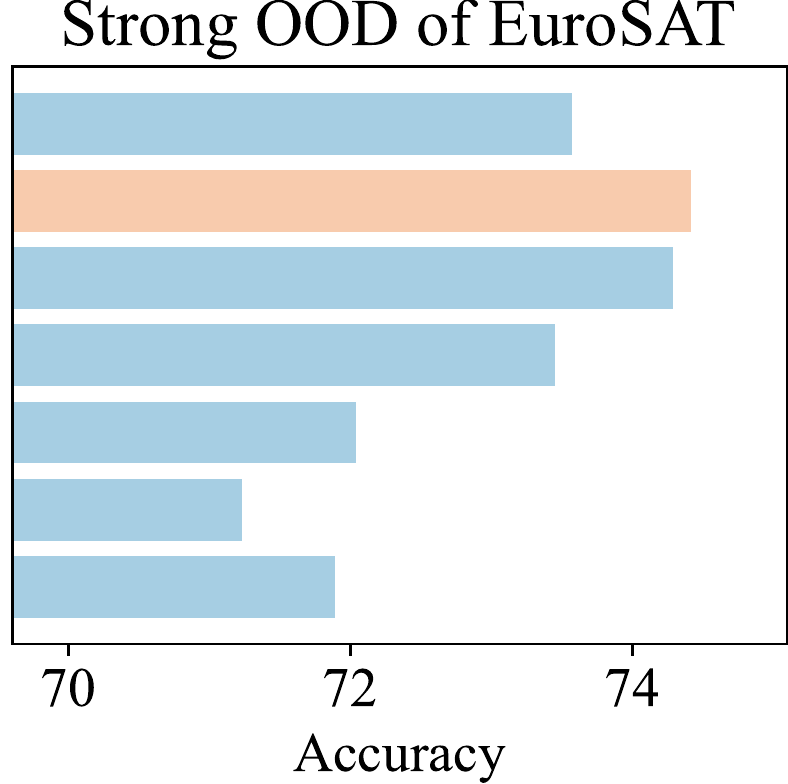} 
	\vspace{-4pt}
  \caption{Few-shot classification accuracy with different hyperparameter $\beta$ for the weak and strong OOD generalization problems under the 5-way 5-shot setting.}
  \label{fig:abeta2}
  \vspace{-6pt}
\end{figure*}

\paragraph{The hyperparameter of $\beta$}
The hyperparameter $\beta$ is introduced in equation~(\ref{eq:rewe}) in Section~\ref{sec:reweight}. The $\beta$ aims to adjust the tasks learning rate and $\beta=0.1$ is used for the weak OOD generalization problem and $\beta=1.5$ is used for the strong OOD generalization problem.
It is because a lower $\beta$ is appropriate for similar tasks and a higher $\beta$ should be used for dissimilar tasks.
More experimental results for $\beta$ are shown in Figures~\ref{fig:abeta1} and~\ref{fig:abeta2}, where ``None'' means the performance of iMAML and the bars with ``\textcolor{red}{red}'' represents the performance of our method (iMAML-\ours).

From Figures~\ref{fig:abeta1} and~\ref{fig:abeta2}, $\beta=0$ damages the performance of vanilla iMAML in the strong OOD generalization problem. 
The main reason is that $\beta$ = 0 tries to pin the backpropagated gradients of each task to be equal at network parameters. 
For the strong OOD generalization problem, this is unreasonable since each task comes from a different distribution, which varies widely from each other.
Moreover, we have two hyperparameters and experiment with a control variable strategy.

\begin{figure*}[t]
	\centering
	\includegraphics[width=0.29\textwidth]{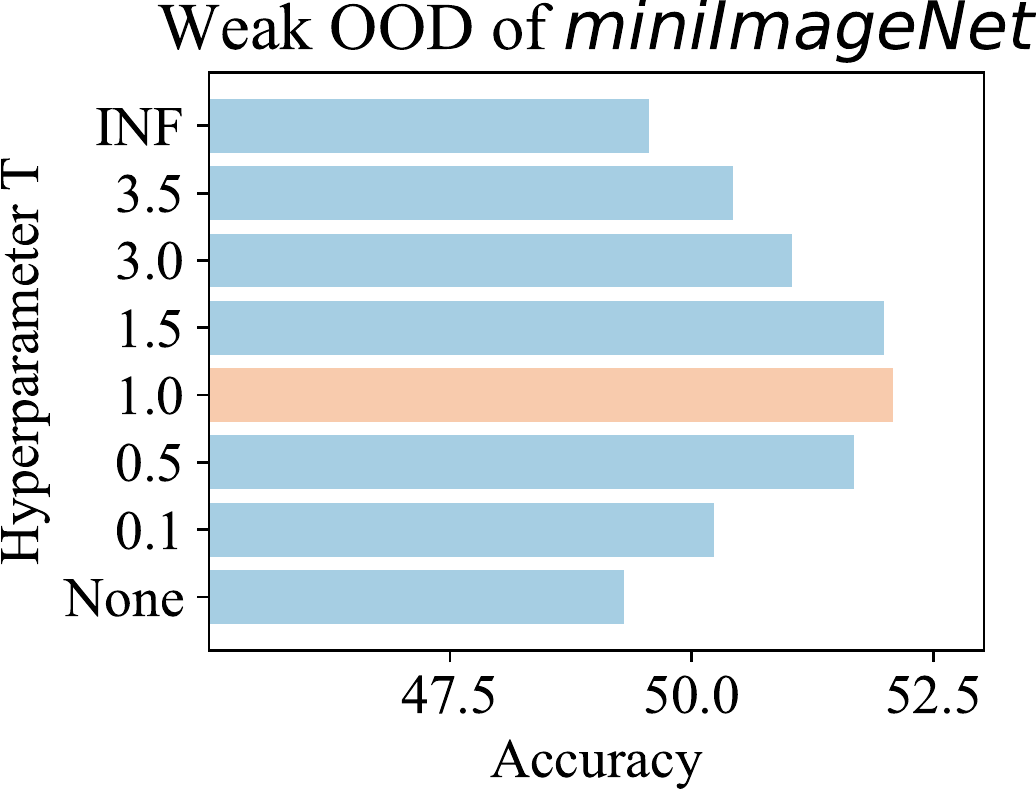} 
    \includegraphics[width=0.22\textwidth]{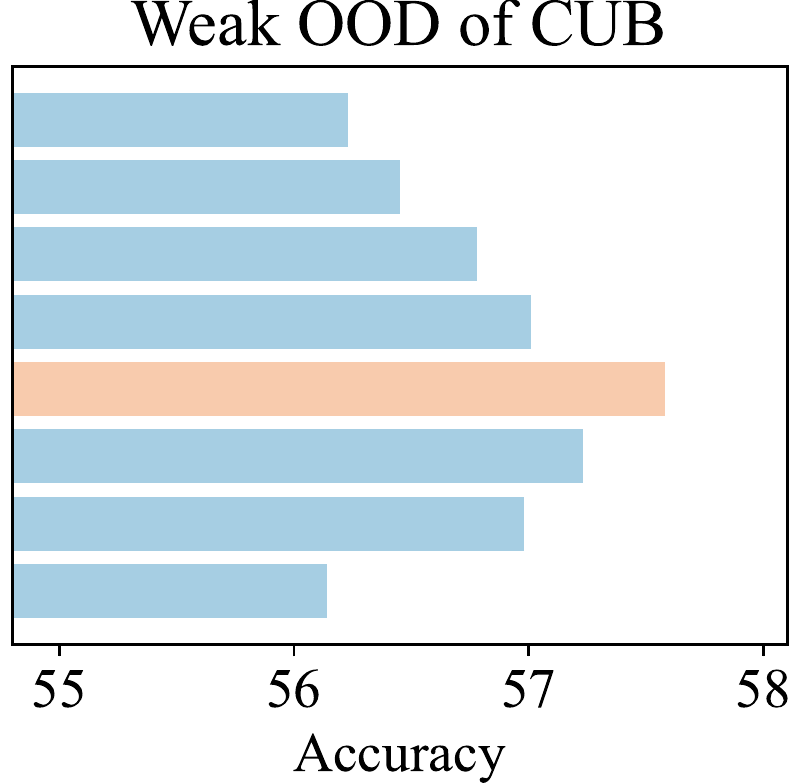}
    \includegraphics[width=0.22\textwidth]{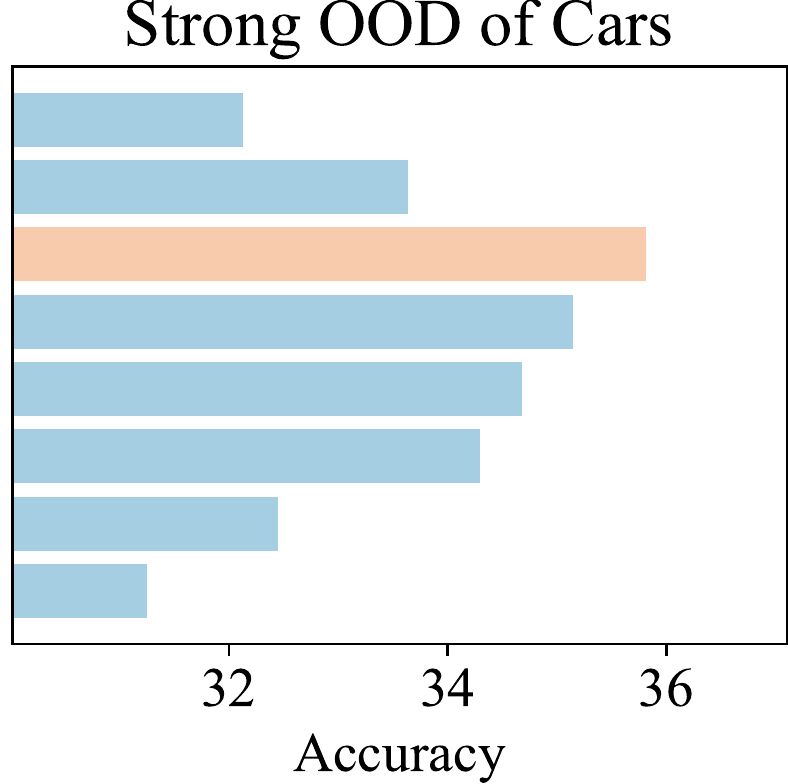}
    \includegraphics[width=0.22\textwidth]{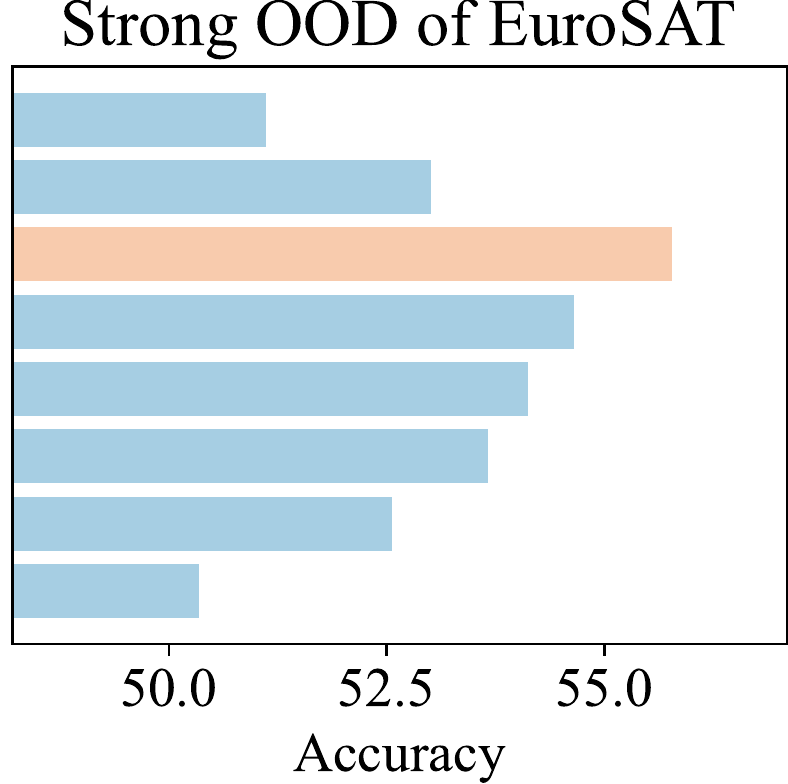} 
	\vspace{-4pt}
  \caption{Few-shot classification accuracy with different hyperparameter $T$ for the weak and strong OOD generalization problems under the 5-way 1-shot setting.}
  \label{fig:aT1}
  \vspace{-6pt}
\end{figure*}

\begin{figure*}[t]
	\centering
	\includegraphics[width=0.29\textwidth]{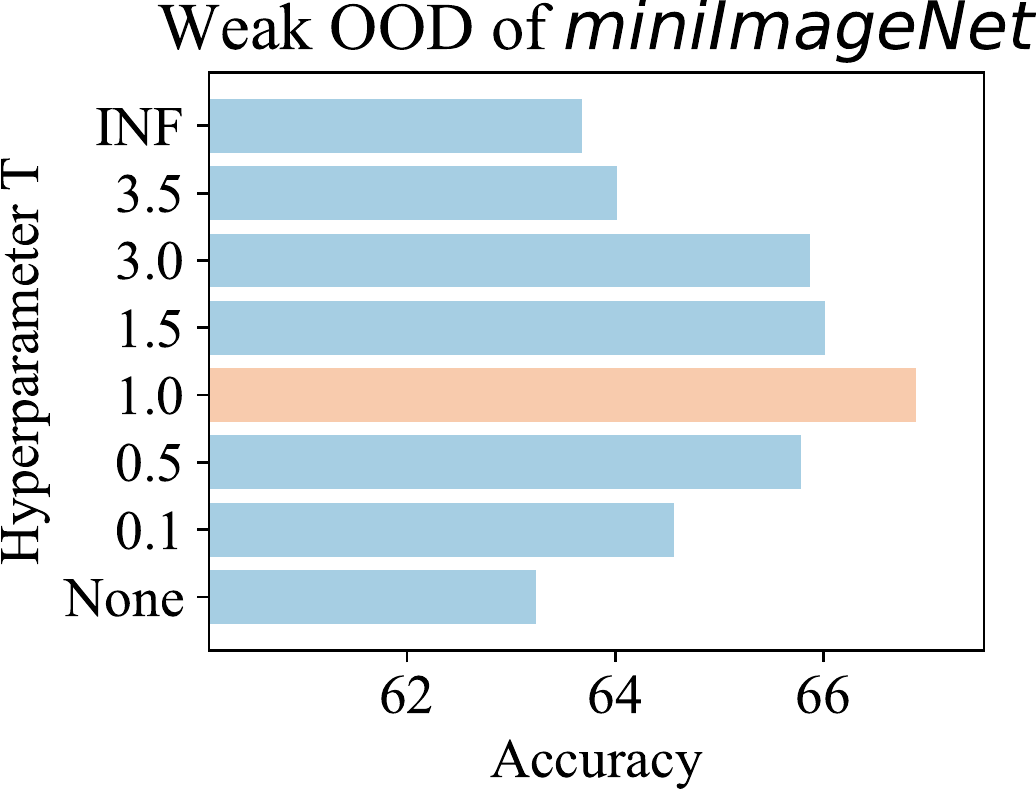} 
    \includegraphics[width=0.22\textwidth]{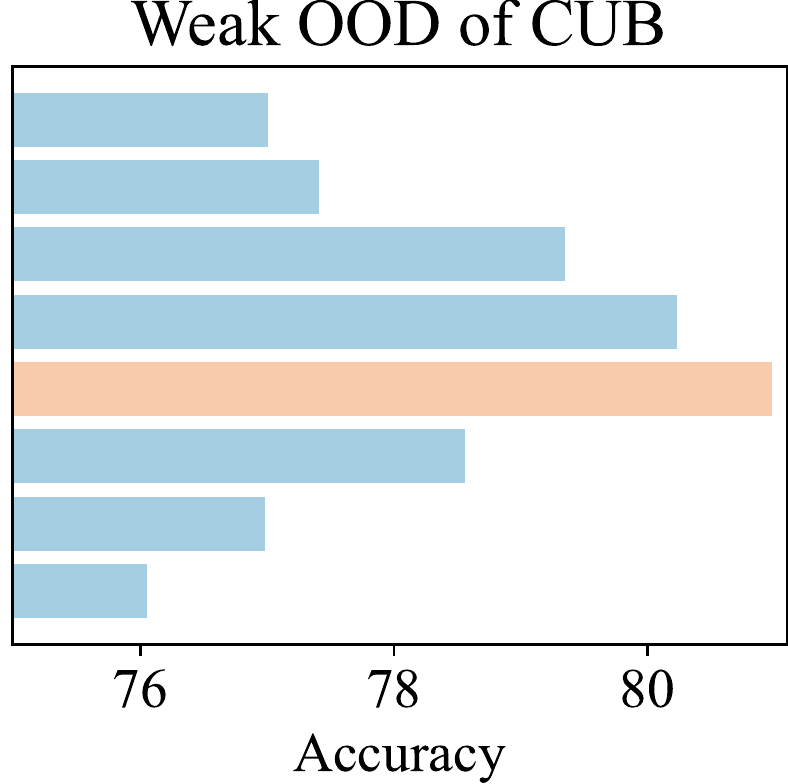}
    \includegraphics[width=0.22\textwidth]{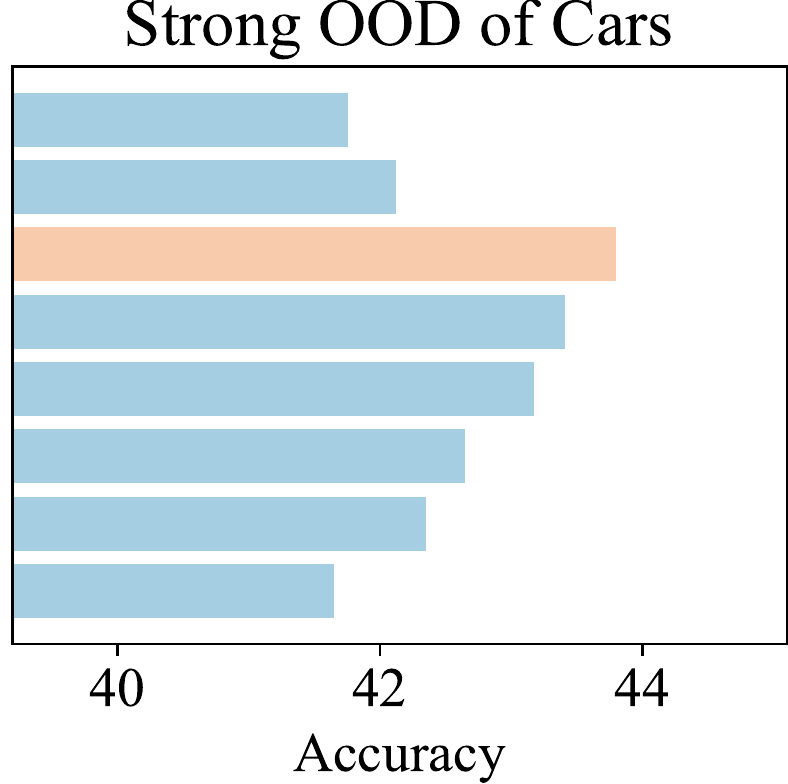}
    \includegraphics[width=0.22\textwidth]{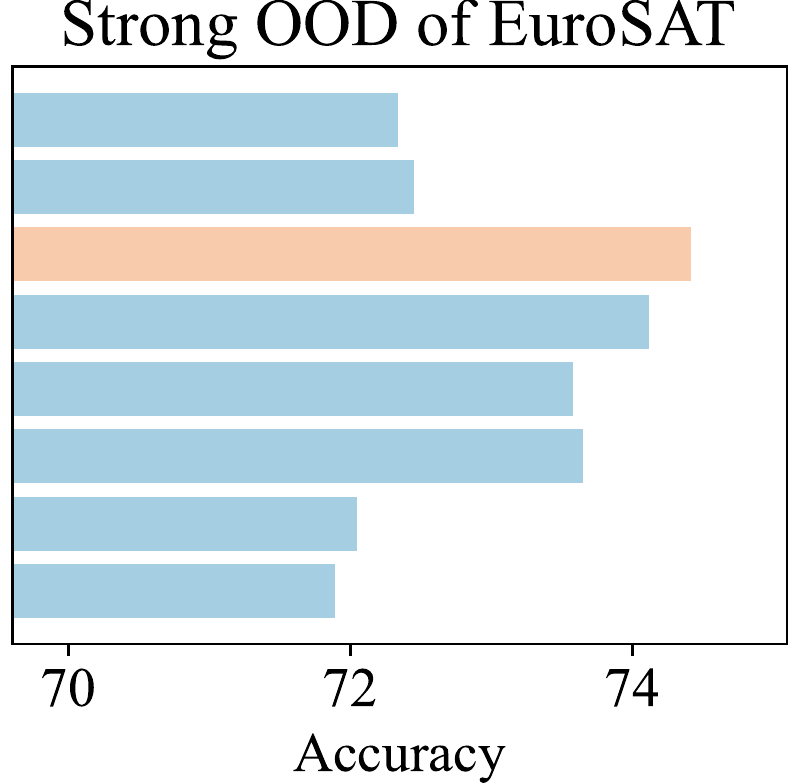} 
	\vspace{-4pt}
  \caption{Few-shot classification accuracy with different hyperparameter $T$ for the weak and strong OOD generalization problems under the 5-way 5-shot setting.}
  \label{fig:aT2}
\end{figure*}

\paragraph{The hyperparameter of temperature $T$}
The hyperparameter of temperature $T$ is used as a ``soft threshold" of information in equation~(\ref{eq:isi}) in Section~\ref{subsec:isi}. When $T$ is small, this means more conservative filtering, \textit{i.e.}, only patches with the least information will be dropped and most shape and texture are preserved. When $T$ goes to infinity, all neurons will be dropped with equal probability, and the whole algorithm becomes regular Dropout~\cite{srivastava2014dropout}. As mentioned above, we have two hyperparameters, so we adopt a strategy of fixing one to experiment with the other. The experimental results are shown in Figure~\ref{fig:aT1} under the 5-way 1-shot and Figure~\ref{fig:aT2} under the 5-way 5-shot setting, where ``None'' means the performance of iMAML, ``INF'' means that it degrades to regular Dropout and the bars with ``\textcolor{red}{red}'' represents the performance of our method (iMAML-\ours). 

From Figures~\ref{fig:aT1} and ~\ref{fig:aT2}, we find that the strong OOD generalization problem uses a larger temperature than the weak OOD generalization problem.
This phenomenon is consistent with our analysis, because the strong OOD generalization problem has greater distribution differences than the weak OOD generalization problem, and more useless information needs to be dropped.

\paragraph{Visualization of the invariant self-information module}
We propose the invariant self-information (ISI) module to avoid some non-causal features to affect the generalization of the neural network on some new tasks from new distributions in Section~\ref{subsec:isi}.
Motivated by~\cite{geirhos2018imagenet,shi2020informative}, we extract the robust shape features as invariant causal features.
To verify the learning of shape features with the ISI module, we visualize gradients of model output using the saliency map.
Specifically, we use SmoothGrad~\cite{smilkov2017smoothgrad} to calculate the saliency map $S(x)$:

\begin{equation}
    S(x) = \frac{1}{n}\sum\limits_{i=1}^{n}\frac{\partial f(x_{i})}{\partial x_{i}},
\end{equation} 

where $x_{i} = x + \delta_{i}$ is original image $x$ with i.i.d. Gaussian noise $\delta_{i}$ and $f(\cdot)$ is the network. 
The experiments are shown in Figure~\ref{fig:sm}, where these original images are randomly sampled from the nine datasets used in our experiments.
We can see that ISI is more human-aligned and sensitive to the shapes of objects. While the Saliency map of the baseline (iMAML) algorithm is filled with noise, which indicates the baseline method is less shape-biased and lacks interpretability.
Although on some difficult examples (the 3-rd, 4-th and 5-th columns), our ISI module also has impressive performance. 

\section{Conclusion}
\label{sec:con}
In this paper, we address the challenging problem of learning a good meta-knowledge of gradient-based meta-learning (GBML) algorithms under the OOD setting. We show that OOD generalization exacerbates the inconsistencies in task-gradient magnitudes and directions. Therefore, we propose RotoGBML, a general framework to dynamically reweight diverse magnitudes to a common scale and rotate conflicting directions close to each other. Moreover, we also design an invariant self-information (ISI) module to extract the invariant causal features. Homogenizing gradients using these causal features provide further guarantees for learning robust meta-knowledge. Finally, we analyze the feasibility of our method from the theoretical and experimental levels. We believe that our work makes a meaningful step toward adjusting task gradients under the OOD problem for GBML.

\bibliography{ref}
\bibliographystyle{IEEEtran}

\onecolumn
\appendix

\section{Theoretical Analysis}
\label{sec:atheory}

\vspace{-5pt}
\subsection{RotoGBML as Stackelberg games}
\label{subsec:agane}
The optimization objective of learning the meta-knowledge in GBML algorithms has two loops: an inner loop and an outer loop. In the outer loop, we simultaneously optimize the network parameters $(\phi, \theta)$ and the module parameters of \ours $(\omega, \gamma)$.
The two optimization objectives can be interpreted as a Stackelberg game~\cite{TannerFiez2019ConvergenceOL}, which is an asymmetric game with players playing alternately. The neural networks $(\phi, \theta)$ are called as the follower to minimize their loss function. The reweighted vector set and rotation matrix set $(\omega, \gamma)$ are named as the leader. Compared to the follower, the leader attempts to minimize their own loss function, but it does so with the advantage of knowing which will be the best response to their movement by the follower. The problem can be written as:
\begin{align}
    & Leader: \ \min\limits_{(\omega, \gamma)} \{\mathcal{L}_{(\omega, \gamma)}(\omega, \gamma, \phi, \theta)| (\phi, \theta)\in \mathop{\arg\min}\limits_{(\phi, \theta)} \sum\nolimits_{i=1}^{N} \mathcal{L}_{i}(\phi, \theta, \omega, \gamma) \},  \\
    & Follower: \ \min\limits_{(\phi, \theta)} \sum\nolimits_{i=1}^{N} \mathcal{L}_{i}(\phi, \theta, \omega, \gamma) .  
\end{align}
The leader knows the global objective while the follower only has access to their own local loss function. In \textit{gradient-play} Stackelberg game, all players perform gradient updates at $s$-th step:
\begin{align}
    (\omega^{s}, \gamma^{s}) &= (\omega^{s-1}, \gamma^{s-1}) - \eta_{(\omega, \gamma)} \nabla_{(\omega, \gamma)} \mathcal{L}_{(\omega^{s-1}, \gamma^{s-1})}(\omega^{s-1}, \gamma^{s-1}, \phi^{s-1}, \theta^{s-1}),   \\
    (\phi^{s}, \theta^{s}) &= (\phi^{s-1}, \theta^{s-1}) - \eta_{(\phi, \theta)} \sum\nolimits_{i=1}^{N} \nabla_{(\phi^{s-1}, \theta^{s-1})} \mathcal{L}_{i}(\phi^{s-1}, \theta^{s-1}, \omega^{s-1}, \gamma^{s-1}),
\end{align}
where $\eta_{(\omega, \gamma)}$ and $\eta_{(\phi, \theta)}$ are the learning rates of the RotoGBML and neural network, respectively. 

Equilibrium is an important concept in game theory, the point at which both players are satisfied with their situation, meaning that there is no action available that can immediately improve any player's score.
If the two players can converge to an equilibrium point, this also guarantees the convergence of the training phase.
We introduce the definition of Stackelberg equilibrium~\cite{fiez2020implicit,javaloy2021rotograd}:

\begin{definition}
    (Differential Stackelberg equilibrium). \ If $\nabla_{(\omega, \gamma)}\mathcal{L}_{(\omega, \gamma)}(\omega, \gamma, r(\omega, \gamma))=0$ and $\nabla_{(\omega, \gamma)}^{2}\mathcal{L}_{(\omega, \gamma)}(\omega, \gamma, r(\omega, \gamma))$ is positive definite, the pair of points $(\phi, \theta)$, $(\omega, \gamma)$ is a differential Stackelberg equilibrium point and $(\phi, \theta)=r(\omega, \gamma)$ is implicitly defined by $\nabla_{(\phi, \theta)}\mathcal{L}_{i}(\phi, \theta, \omega, \gamma)=0$.
\label{de:dsq}
\end{definition}

\vspace{-8pt}
When the players manage to reach such an equilibrium point, both of them are at a local optimum. Therefore, here we also provide theoretical convergence guarantees for an equilibrium point.
\begin{proposition}
    In the given setting, if the leader's learning rate $\eta_{(\omega, \gamma)}$ goes to zero at a faster rate than the follower's $\eta_{(\phi, \theta)}$, that is, they will asymptotically converge to a differential Stackelberg equilibrium point almost surely.
\label{pro:con}    
\end{proposition}
As long as the follower learns faster than the leader, they will end up in a situation where both are satisfied. In our experiments, the learning rate of the follower is $1e-3$ and that of the leader is $5e-4$.

\subsection{Proof of Theorem 4.1}
\label{subsec:proof1}
In our main paper, we propose the Theorem 4.1 to bound the difference of task gradients $d_{ij}=\|g_{i}-g_{j}\| \le 4 \eta_{base} GL \|\mathbb{P}_{i}(\mathcal{T})-\mathbb{P}_{j}(\mathcal{T})\|_{TV}$ by using the total variation distance (TVD) based on an assumption of $G$ and $L$. Motivated by previous work~\cite{fallah2021generalization}, the assumption is defined as follows:

\begin{assumption}
   For any $x\in X$, the loss function $\mathcal{L}(\cdot, x)$ is twice continuously differentiable. Furthermore, we assume it satisfies the following properties for any $\psi=(\phi,\theta), \bar{\psi}=(\bar{\phi}, \bar{\theta})$:
   \begin{itemize}
       \item The gradient norm is uniformly bounded by $G$ over $\psi$, \textit{i.e.}, $\|\nabla \mathcal{L}(\psi, x)\| \le G$.
       \item The loss is $L$-smooth, \textit{i.e.}, $\|\nabla \mathcal{L}(\psi, x) - \nabla \mathcal{L} (\bar{\psi}, x) \| \le L \|\psi- \bar{\psi}\|$.
   \end{itemize}
   \label{assu:l}
\end{assumption}

\vspace{-10pt}
\begin{proof}
$\mathcal{T}_{i}=(\mathcal{S}_{i}, \mathcal{Q}_{i})\sim \mathbb{P}_{i}(\mathcal{T})$ and $\mathcal{T}_{j}=(\mathcal{S}_{j}, \mathcal{Q}_{j})\sim \mathbb{P}_{j}(\mathcal{T})$ with $i \neq j$ are randomly sampled from the different distributions. The difference of task gradients is definited as follows:
\begin{equation}
\begin{aligned}
    d_{ij} &= \|\nabla_{\psi}\mathcal{L}_{i}(\psi_{i}^{\tau}(x_{i}^{q}), y_{i}^{q}) - \nabla_{\psi}\mathcal{L}_{j}(\psi_{j}^{\tau}(x_{j}^{q}), y_{j}^{q}) \|,  \\
    &= \|\nabla_{\psi}\mathcal{L}_{i}(\psi- in_{i}, x_{i}^{q}, y_{i}^{q}) - \nabla_{\psi}\mathcal{L}_{j}(\psi - in_{j}, x_{j}^{q}, y_{j}^{q})\|,    \\
    & = \|\mathbb{E}_{\{(x_{i}^{k}, y_{i}^{k})\sim\mathcal{Q}_{i}\sim\mathbb{P}_{i}(\mathcal{T})\}_{k=1}^{n_{q}}} \nabla_{\psi} \mathcal{L}_{i}(\psi- in_{i}, x_{i}^{k}, y_{i}^{k})  \\
    & \, \, \, \, \, \, \, - \mathbb{E}_{\{(x_{j}^{k}, y_{j}^{k})\sim\mathcal{Q}_{j}\sim\mathbb{P}_{j}(\mathcal{T})\}_{k=1}^{n_{q}}} \nabla_{\psi} \mathcal{L}_{j}(\psi- in_{j}, x_{j}^{k}, y_{j}^{k}) \|,   \\ 
    & \mathrm{where} \ in_{i} = \eta_{base} \mathbb{E}_{\{(x_{i}^{k}, y_{i}^{k})\sim\mathcal{S}_{i}\sim\mathbb{P}_{i}(\mathcal{T})\}_{k=1}^{n_{s}}} \nabla_{\psi_{i}^{\tau}}\mathcal{L}_{i}(\psi_{i}^{\tau-1}(x_{i}^{k}), y_{i}^{k}),  \\
    & \, \, \, \, \, \, \, \, \, \, \, \, \, \, \, \, \, \, in_{j} = \eta_{base} \mathbb{E}_{\{(x_{j}^{k}, y_{j}^{k})\sim\mathcal{S}_{j}\sim\mathbb{P}_{j}(\mathcal{T})\}_{k=1}^{n_{s}}} \nabla_{\psi_{j}^{\tau}}\mathcal{L}_{j}(\psi_{j}^{\tau-1}(x_{j}^{k}), y_{j}^{k}), 
\label{eq:adij}
\end{aligned}
\end{equation}
where $n_{s}$ and $n_{q}$ are the number of the support set and query set examples, respectively.
$g_{i} = \nabla_{\psi} \mathcal{L}_{i}(\psi_{i}^{\tau}(x_{i}^{q}), y_{i}^{q})$ and $g_{j} = \nabla_{\psi} \mathcal{L}_{j}(\psi_{j}^{\tau}(x_{j}^{q}), y_{j}^{q})$ are the task gradients for $\mathcal{T}_{i}$ and $\mathcal{T}_{j}$ using query set in the outer loop, respectively.
Hence, with probability $\binom{n_{q}}{t} (\|\mathbb{P}_{i}(\mathcal{T}) - \mathbb{P}_{j}(\mathcal{T}) \|_{TV})^t (1-\| \mathbb{P}_{i}(\mathcal{T}) - \mathbb{P}_{j}(\mathcal{T}) \|_{TV})^{n_{q}-t}$, we have $x_{i}^{k} \neq x_{j}^{k}$ for $t$ choices of $k$ (out of $1,...,n_{q}$). In addition, based on the Assumption~\ref{assu:l}, the equation~(\ref{eq:adij}) is bounded as follows:
\begin{small}
\begin{equation}
    \begin{aligned}
    d_{ij} &= \|\mathbb{E}_{\{(x_{i}^{k}, y_{i}^{k})\sim\mathcal{Q}_{i}\sim\mathbb{P}_{i}(\mathcal{T})\}_{k=1}^{n_{q}}} \nabla_{\psi} \mathcal{L}_{i}(\psi- in_{i}, x_{i}^{k}, y_{i}^{k}) - \mathbb{E}_{\{(x_{j}^{k}, y_{j}^{k})\sim\mathcal{Q}_{j}\sim\mathbb{P}_{j}(\mathcal{T})\}_{k=1}^{n_{q}}} \nabla_{\psi} \mathcal{L}_{j}(\psi- in_{j}, x_{j}^{k}, y_{j}^{k}) \|,   \\
    & \le 2 \eta_{base} L \|\mathbb{E}_{\{(x_{i}^{k}, y_{i}^{k})\sim\mathcal{Q}_{i}\sim\mathbb{P}_{i}(\mathcal{T})\}_{k=1}^{n_{q}}}\nabla_{\psi} \mathcal{L}_{i}(\psi_{i}^{\tau}(x_{i}^{k}), y_{i}^{k}) - \mathbb{E}_{\{(x_{j}^{k}, y_{j}^{k})\sim\mathcal{Q}_{j}\sim\mathbb{P}_{j}(\mathcal{T})\}_{k=1}^{n_{q}}}\nabla_{\psi} \mathcal{L}_{j}(\psi_{j}^{\tau}(x_{j}^{k}), y_{j}^{k}) \|, \\
    & \le 4 \eta_{base} GL \frac{t}{K}. 
    \end{aligned}
\end{equation}
\end{small}
As a result, we have
\begin{small}
\begin{equation}
    \begin{aligned}
    d_{ij} & = \|\mathbb{E}_{\{(x_{i}^{k}, y_{i}^{k})\sim\mathcal{Q}_{i}\sim\mathbb{P}_{i}(\mathcal{T})\}_{k=1}^{n_{q}}} \nabla_{\psi} \mathcal{L}_{i}(\psi- in_{i}, x_{i}^{k}, y_{i}^{k}) - \mathbb{E}_{\{(x_{j}^{k}, y_{j}^{k})\sim\mathcal{Q}_{j}\sim\mathbb{P}_{j}(\mathcal{T})\}_{k=1}^{n_{q}}} \nabla_{\psi} \mathcal{L}_{j}(\psi- in_{j}, x_{j}^{k}, y_{j}^{k}) \|,  \\
    & \leq \sum_{t=0}^K \binom{K}{t} (\| \mathbb{P}_{i}(\mathcal{T}) - \mathbb{P}_{j}(\mathcal{T}) \|_{TV})^t (1-\| \mathbb{P}_{i}(\mathcal{T}) - \mathbb{P}_{j}(\mathcal{T}) \|_{TV})^{K-t} \cdot 4 \eta_{base} GL \frac{t}{K}, \\
    & = 4 \eta_{base} GL (\|\mathbb{P}_{i}(\mathcal{T}) - \mathbb{P}_{j}(\mathcal{T}) \|_{TV}) \sum_{t=0}^K \frac{t}{K} \binom{K}{t} (\| \mathbb{P}_{i}(\mathcal{T}) - \mathbb{P}_{j}(\mathcal{T}) \|_{TV})^{t-1} (1-\|\mathbb{P}_{i}(\mathcal{T}) - \mathbb{P}_{j}(\mathcal{T}) \|_{TV})^{K-t},  \\
    & = 4 \eta_{base} GL (\|\mathbb{P}_{i}(\mathcal{T}) - \mathbb{P}_{j}(\mathcal{T}) \|_{TV}),	  \\
    \end{aligned}
\end{equation}
\end{small}
where the last equality follows from the fact that
\begin{small}
\begin{equation}
    \begin{aligned}
    &\frac{t}{K} \binom{K}{t} (\| \mathbb{P}_{i}(\mathcal{T}) - \mathbb{P}_{j}(\mathcal{T}) \|_{TV})^{t-1} (1-\| \mathbb{P}_{i}(\mathcal{T}) - \mathbb{P}_{j}(\mathcal{T}) \|_{TV})^{K-t}  \\
    & \, \, = \binom{K-1}{t-1} (\| \mathbb{P}_{i}(\mathcal{T}) - \mathbb{P}_{j}(\mathcal{T}) \|_{TV})^{t-1} (1-\| \mathbb{P}_{i}(\mathcal{T}) - \mathbb{P}_{j}(\mathcal{T}) \|_{TV})^{K-1-(t-1)}.
    \end{aligned}
\end{equation}
\end{small}
\vspace{1.2pt}
Therefore, we can prove the conclusion of $d_{ij}=\|g_{i}-g_{j}\| \le 4 \eta_{base} GL \|\mathbb{P}_{i}(\mathcal{T})-\mathbb{P}_{j}(\mathcal{T})\|_{TV}$.
\end{proof}

\end{document}